%% file: paper.tex
\documentclass{article}

\usepackage{arxiv}

\usepackage{graphicx}
\usepackage[utf8]{inputenc} 
\usepackage[T1]{fontenc}    
\usepackage{lineno,hyperref}       
\usepackage{url}            
\usepackage{booktabs}       
\usepackage{amsmath,amsfonts}       
\usepackage{nicefrac}       
\usepackage{microtype}      
\usepackage{lipsum}
\usepackage{xcolor, colortbl}
\usepackage{bm}
\usepackage{hhline}
\usepackage{chngpage}

\def\x{{\bm x}}

\title{A Nature-Inspired Feature Selection Approach based on Hypercomplex Information}

\author{
  Gustavo H. de Rosa, João P. Papa \\
  Department of Computing \\
  São Paulo State University \\
  Bauru, São Paulo - Brazil \\
  \texttt{gustavo.rosa@unesp.br, joao.papa@unesp.br} \\
  \And
  Xin-She Yang \\
  School of Science and Technology \\
  Middlesex University \\
  London, United Kingdom \\
  \texttt{x.yang@mdx.ac.uk} \\
}
\begin{document}

\maketitle

\begin{abstract}
Feature selection for a given model can be transformed into an optimization task. The essential idea behind it is to find the most suitable subset of features according to some criterion. Nature-inspired optimization can mitigate this problem by producing compelling yet straightforward solutions when dealing with complicated fitness functions. Additionally, new mathematical representations, such as quaternions and octonions, are being used to handle higher-dimensional spaces. In this context, we are introducing a meta-heuristic optimization framework in a hypercomplex-based feature selection, where hypercomplex numbers are mapped to real-valued solutions and then transferred onto a boolean hypercube by a sigmoid function. The intended hypercomplex feature selection is tested for several meta-heuristic algorithms and hypercomplex representations, achieving results comparable to some state-of-the-art approaches. The good results achieved by the proposed approach make it a promising tool amongst feature selection research.
\end{abstract}


\input{introduction.tex}
\input{hypercomplex.tex}
\input{feature-selection.tex}
\input{methodology.tex}

\input{results.tex}
\input{conclusion.tex}

\section*{Acknowledgments}
The authors appreciate S\~ao Paulo Research Foundation (FAPESP) grants \#2013/07375-0, \#2014/12236-1, \#2016/19403-6, \#2017/02286-0, \#2017/25908-6, \#2018/21934-5 and \#2019/02205-5, and CNPq grants 307066/2017-7 and 427968/2018-6.

\bibliographystyle{unsrt}  
\bibliography{references}

\end{document}

%% file: introduction.tex
\section{Introduction}
\label{s.introduction}

Optimization techniques became more and more popular in the last few years. Beneficial in numerous applications, ranging from engineering~\cite{Yang:10, OhASC:17}, medicine~\cite{Klein:07,Dey:14} to machine learning fine-tuning~\cite{RosaCIARP:15,PapaGECCO:15,PapaASC:16,PapaJoCS:15}, they provide suitable solutions and virtually none human interaction with the modeling process, leaving the burden of choosing parameters to the model itself. In this context, most of the obstacles described by non-convex mathematical functions~\cite{Bertsekas:99} requires more robust optimization approaches rather than conventional optimization methods.

Meta-heuristics algorithms, usually referred to as nature-inspired, or even to swarm- or evolutionary-based algorithms, gained great attention in the last years, attempting to solve optimization problems in a more appealing way than traditional methods. These so-called nature techniques work without derivatives, thus being suitable for problems with high dimensional spaces. Even though they provide outstanding results in different applications, they can still get trapped into local optimal points. Thus, an important question is how to run these algorithms in the case of complex objective functions. One can refer to hybrid variants~\cite{HuTEC:10}, aging mechanisms~\cite{ChenTEC:13}, and fitness landscape analysis~\cite{Pitzer:12} as some distinct strategies used to deal with this issue.

As mentioned above, the problem of selecting possible parameters can be solved as an optimization problem, where a subset of parameters or features can be used to calculate the value of a fitness function. This is similar to feature selection, and it is usually classified into two divisions: (i) wrapper approaches~\cite{Kohavi:97}, and (ii) filter-based~\cite{Sanchez:07}. The former methods use the output of some classifier (e.g., classification accuracy) to control the optimization method. Conversely, filter-based ones do not consider this information.

One can presume that feature selection is a straightforward solution that automatizes the choice of parameters. However, it is still necessary to select an appropriate fitness function, which is regularly correlated to the problem's nature. Also, most machine learning problems deal with high-dimensional data, thus amplifying the problem of exploring the search space. An intriguing way to tackle this obstacle is to use a more complex representation of the search space, the so-called hypercomplex search space. The goal behind handling hypercomplex spaces is based on the possibility of having more natural fitness landscapes, although it has not been mathematically proved yet. Nevertheless, the results achieved previously sustain such a hypothesis~\cite{FisterESA:13,FisterIEEECEC:15,PapaANNPR:16,PapaASC:17}.

Normalized quaternions, also known as versors, are broadly used to describe the orientation of objects in three-dimensional spaces, being extremely efficient in performing rotations in such spaces~\cite{Hart:94}. Another intriguing addition of quaternions are the octonions, comprised of eight dimensions~\cite{Graves:45}. Even though they are not well known in the literature, they have compelling traits that make them suitable for special relativity and quantum mechanics, among other research specialties~\cite{DeLeo:96, Finkelstein:62}. However, to the best of our knowledge, they have not been used to embed search spaces in meta-heuristic feature selection so far.

This study considers $8$ meta-heuristic techniques, among with their quaternion- and octonion-based versions, validated under $20$ different datasets, proving the robustness of quaternionic and octatonic representations for hypercomplex-embedded search spaces. Therefore, we believe this paper can serve as a foundation for prospective research regarding hypercomplex representations in the context of meta-heuristic-based feature selection.

The rest of this paper is organized as follows. Sections~\ref{s.hypercomplex} and~\ref{s.feature_selection} present the theoretical background related to hypercomplex-based spaces (quaternions and octonions), and the proposed approach for hypercomplex-based feature selection, respectively. Section~\ref{s.methodology} discusses the methodology and the computational setup adopted in this paper, while Section~\ref{s.results} presents the numerical results. Finally, Section~\ref{s.conclusion} states conclusions and future works.

%% file: hypercomplex.tex
\section{Hypercomplex-based Spaces}
\label{s.hypercomplex}

\subsection{Complex Numbers}
\label{ss.complex}

The following problems can be solved with the modern methods of numerical analysis:

\begin{equation}
\label{e.complex}
    x^2+1=0,
\end{equation}

in spite of the fact that $x^2=-1$ cannot be a rational solution as any number square root must be positive, $ x\in\Re$.

The problem (\ref{e.complex}) can be solved using the imaginary representation:

\begin{equation}
    i^2=-1,
\end{equation}
although this may not appear to be logically correct. The imaginary numbers assemble a structure called \emph{complex numbers}, which is formed by real and imaginary terms, as follows:

\begin{equation}
    c=h_0+h_1i,
\end{equation}
where $h_0,h_1\in\Re$ and $i^2=-1$. One can perceive that it is feasible to obtain a real number by using $h_1=0$, or even an imaginary number by placing $h_0=0$. Thus, the complex numbers deal with the generalization of both real and imaginary numbers.

\begin{figure}[!htb]
\centering
\includegraphics[scale=0.315]{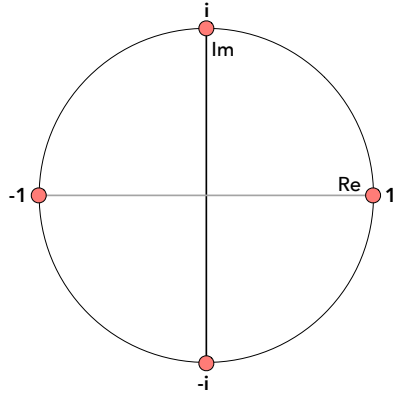}
\caption{\small Representation of a complex plane, which is used to map complex numbers onto a two-dimensional space.}
\label{f.complex_plane}
\end{figure}

One striking operation that performs positively well in a two-dimensional space is the rotation of complex numbers. Firstly, let us map a complex number on a two-dimensional grid, called \emph{complex plane}, where the horizontal axis holds the real part mapping (\textbf{Re}), and the vertical axis is accountable for the imaginary part (\textbf{Im}). This description is depicted by Figure~\ref{f.complex_plane}.

One can see that we need to multiply a complex number by $i$ for each 90-degree rotation in the complex plane. To clarify this, let us consider a random point denoted by $r=i+1$. Also, let $x$ be the result of the multiplication of $r$ by $i$, as follows:

\begin{equation}
    x = ri = i+i^2 = -1+i.
\end{equation}

Now, we can obtain a singular $y$ point by multiplying again $x$ by $i$:

\begin{equation}
    y = xi = -i+i^2 = -1-i.
\end{equation}

Moreover, if we multiply the result $y$ by $i$, a $w$ point can be achieved as follows:

\begin{equation}
    w = yi = -i-i^2 = 1-i.
\end{equation}

\begin{figure}[!htb]
\centering
\includegraphics[scale=0.315]{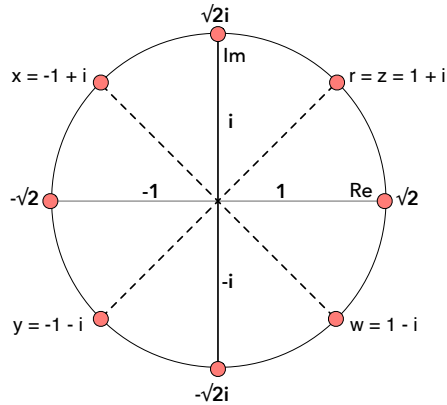}
\caption{\small Representation of complex numbers' rotation throughout the complex plane.}
\label{f.rotation}
\end{figure}

Finally, by multiplying $w$ with $i$, we conclude:

\begin{equation}
    z = wi = i-i^2 = 1+i,
\end{equation}
where $z$ is the same first defined position, i.e., $r=z$. Figure~\ref{f.rotation} illustrates the above calculations.

\subsection{Hypercomplex Numbers}
\label{ss.hypercomplex}

In a particular behavior, we can extend the idea of complex numbers by adding new imaginary terms, producing the so-called hypercomplex numbers. This concept also allows rotations to be performed in higher-dimensional complex spaces. In this work, we consider two traditional hypercomplex representations: quaternions and octonions.

\subsubsection{Quaternions}
\label{sss.quaternions}

A quaternion $q$ is a hypercomplex number, composed of real and complex parts, being $q=h_0+h_1i+h_2j+h_3k$, where $h_0,h_1,h_2,h_3\in\Re$ and $i,j,k$ are imaginary numbers (also known as ``fundamental quaternions units"). This assumption is hold by the following set of equations:

\begin{equation}
ij=k,
\end{equation}
\begin{equation}
jk=i,
\end{equation}
\begin{equation}
ki=j,
\end{equation}
\begin{equation}
ji=-k,
\end{equation}
\begin{equation}
kj=-i,
\end{equation}
\begin{equation}
ik=-j,
\end{equation}
and
\begin{equation}
i^2=j^2=k^2=-1.
\end{equation}

Essentially, a quaternion $q$ is a four-dimensional space representation over the real numbers, i.e., $\Re^4$. 

Given two arbitrary quaternions $q_1=g_0+g_1i+g_2j+g_3k$ and $q_2=h_0+h_1i+h_2j+h_3k$, the quaternion algebra defines a set of main operations~\cite{Eberly:02}. The addition operation, for instance, can be defined as follows:

\begin{align}
q_1+q_2&=(g_0+g_1i+g_2j+g_3k)+(h_0+h_1i+h_2j+h_3k)\\\nonumber
       &=(g_0+h_0)+(g_1+h_1)i+(g_2+h_2)j+(g_3+h_3)k,
\end{align}
while the subtraction is defined as follows:

\begin{align}\nonumber
q_1-q_2&=(g_0+g_1i+g_2j+g_3k)-(h_0+h_1i+h_2j+h_3k)\\
       &=(g_0-h_0)+(g_1-h_1)i+(g_2-h_2)j+(g_3-h_3)k.
\end{align}

Morever, Fister et al.~\cite{FisterESA:13,FisterIEEECEC:15} introduced two other operations, $q_{\text{rand}}$ and $q_{\text{zero}}$. The former initializes a given quaternion with values drawn from a Gaussian distribution $\cal N$, and is defined as follows:

\begin{equation}
    q_{\text{rand}}()=\{g_i={\cal N}(0,1) \mid i\in\{0,1,2,3\}\}.
\end{equation}

The latter equation initializes a quaternion with zero values, as follows:

\begin{equation}
    q_{\text{zero}}()=\{g_i=0 \mid i\in\{0,1,2,3\}\}.
\end{equation}

\subsubsection{Octonions}
\label{sss.octonions}

Octonions are a natural extension of quaternions and were discovered autonomously by John T. Graves and Arthur Cayley around 1843. An octonion is composed of seven complex parts and one real-valued term, being defined as follows:

\begin{equation}
    o=h_0e_0+h_1e_1+h_2e_2+\ldots+h_7e_7, 
\end{equation}
where $h_i\in\Re$ and $e_i$ are the imaginary numbers, $i=0,\ldots,7$. Commonly, $e_0=1$ is used in order to obtain the real-valued term of the octonion.

The addition, subtraction, and norm equations are computed likewise to the quaternions' formulae, giving us a clear implementation framework in order to manipulate several hypercomplex representations.

%% file: feature-selection.tex
\section{Feature Selection}
\label{s.feature_selection}

This section outlines the proposed method for meta-heuristic-based feature selection. One can understand the feature selection process as a method that decides whether a feature should be selected or not (boolean) in order to solve a given problem. As traditional optimization algorithms use a continuous-valued search space, we need to shape the search space into an $n$-dimensional binary structure, where solutions are selected across the edges of a hypercube. Furthermore, as our problem is to select or not a feature, each solution individual is now an $n$-dimensional binary array, where each dimension corresponds to a specific feature and the values $1$ and $0$ indicate whether this feature will or will not be part of the new set.

Concerning conventional optimization algorithms, the solutions are found upon continuous-valued positions of the search space. In order to accomplish this binary-valued individual, one can restrain the new solutions to binary values only:

\begin{equation}
\label{e.transfer1a}
S(x_i^j) = \frac{1}{1+e^{-x_i^j}},
\end{equation}

\begin{eqnarray}
\label{e.transfer1b}
x_i^j & = & \left\{ \begin{array}{ll}
  1\quad & \mbox{if $S(x_i^j)>\alpha$,} \\
0& \mbox{otherwise}
\end{array}\right.
\end{eqnarray}
in which $\alpha\sim U(0,1)$, and $\x \in \Re$ stands for a possible solution.

\begin{figure}[!htb]
\centering
\includegraphics[scale=0.6]{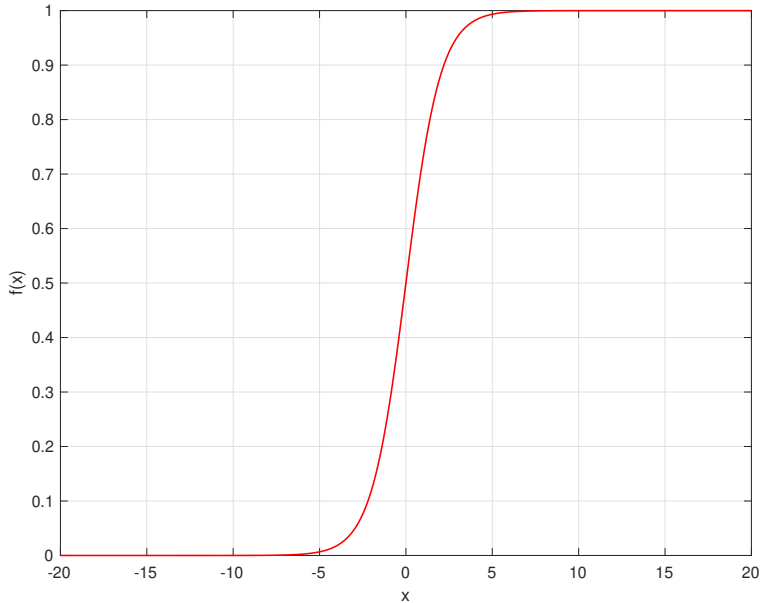}
\caption{\small Sigmoid transfer function $f(x) = \frac{1}{1 + e^{-x}}$ bounded in $[-20, 20]$.}
\label{f.transfer_function}
\end{figure}

Equation~\ref{e.transfer1a} represents the \emph{transfer function}, which maps real-valued solutions into binary-valued ones. Note that any transfer functions can be used to fulfill this purpose. In this work, we are using a sigmoid function (Equation~\ref{e.transfer1a}), which is illustrated by Figure~\ref{f.transfer_function} to map bounded real-valued solutions\footnote{Each real-valued solution} is bounded within the interval $[-20, 20]$. and generate a probability. Further, the mapped value is compared against a uniform distribution sampling in order to obtain the binary output (Equation~\ref{e.transfer1b}).

\subsection{Hypercomplex Feature Selection}

A hypercomplex-based feature selection strategy does not deviate too much from the regular method. One can encode the common search space into a higher-dimensional space, by applying the power of quaternions or octonions. When conducting the meta-heuristic algorithm through the hypercomplex space towards a feasible solution, a crucial operator that needs to be defined is the $p$-norm, which is responsible for mapping hyper-complex values to real numbers. Let $q$ be a hypercomplex number with real coefficients $\left\{h_d\right\}_{d=1}^{D}$, one can compute the Minkowski $p$-norm as follows:

\begin{equation}
\label{e.norm}
    \|q\|_p = \left( \sum_{d=1}^{D} |h_{d}|^p \right)^{1/p}
\end{equation}
where $D$ is the number of dimensions of the space ($2$ for complex numbers, $4$ for quaternions and $8$ for octonions, for instance) and $p \geq 1$. Common values for the latter variable are $1$ or $2$ for the Taxicab and Euclidean\footnote{In this work, we opted to use the Euclidean Norm as mapping function.} norms, respectively. Hence, one can see the $p$-norm as a generalization of such distance operators.

Prior to the transfer function activation, there is an additional equation, called the Span function, which is responsible for mapping the norm's output between the lower and upper bounds, as follows:

\begin{equation}
\label{e.span}
	q_{span}() = (b_u - b_l) \frac{\|q\|_p}{D^{1/p}} + b_l,
\end{equation}
where $b_l$ and $b_u$ stands for the lower and upper bounds, respectively.

\begin{figure}[!htb]
\centering
\includegraphics[scale=0.6]{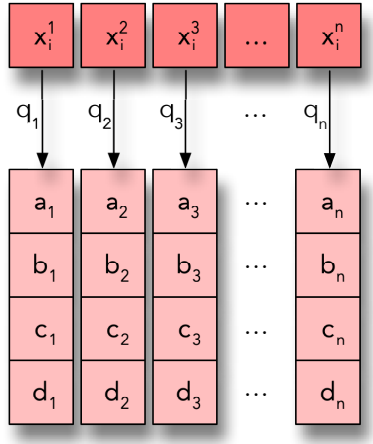}
\caption{\small Quaternionic hypercomplex encoding of a solution vector $\x$, such that $x_i^j$ stands for the $i$-th component of the hypercomplex number for the $j$-th decision variable.}
\label{f.hypercomplex}
\end{figure}

Figure~\ref{f.hypercomplex} illustrates an encoding of a solution vector $\x$ into a quaternionic space, where $x_i^j$ depicts the $i$-th component of the hypercomplex number for the $j$-th decision variable. The same approach can be applied to octatonic spaces by extending the quaternion $q$ (four components) to an octonion $o$ (eight components).

%% file: methodology.tex
\section{Methodology and Setup}
\label{s.methodology}

The idea behind this work is to model the task of selecting the most suitable features for a given problem through a meta-heuristic optimization process. As stated in Section~\ref{s.introduction}, feature selection stands for a proper selection of features, reducing a particular problem's dimensionality and usually enhancing its performance. Also, as the proposed approach is a wrapper-based one, there is a need to define an objective function that will conduct the optimization process. Therefore, the proposed approach aims at selecting the subset of features that minimize the classification error (maximize the classification accuracy) of a given supervised classifier over a validation set. Although any supervised pattern recognition classifier could be applied, we opted to use the Optimum-Path Forest (OPF)~\cite{PapaIJIST:09,PapaPR:12} since it is parameterless and has a fast training procedure. Essentially, the OPF encodes each dataset's sample as a node in a graph, whose connections are defined by an adjacency relation. Its learning process aims at finding prime samples called \emph{prototypes} and trying to conquer the remaining samples by offering them optimum-paths according to a path-cost function. In the end, optimum-path trees are achieved, each one rooted at a different prototype node.

\begin{sloppypar}
Given a set of hypercomplex candidate solutions $\{\bm{s}\}_{i=1}^{S}$, where $\bm{s} = \{s_1, s_2, \ldots, s_j\} \mid j \in \{1, 2, \ldots, N\}$, such that $S$ stands for meta-heuristic candidates and $N$ for the number of decision variables (number of the problem's features depicted in Table~\ref{t.datasets}), we wish to learn the best set of features $F^\star$. Namely, we want to solve the following optimization problem:    
\end{sloppypar}

\begin{equation}
\begin{aligned}
  F^\star =& \left\{ f_i(\bm{s}_i, F) \;\middle|\; i \in \left[1, S\right] \right\}, \\
  & \text{st. } 1 \leq F \leq N.
  \label{e.p_optimization}
\end{aligned}
\end{equation}
where $f_i(\bm{s}_i, F)$ stands for the fitness function (OPF accuracy over validation set) of candidate $i$ based on its binary solution $\bm{s}_i$, which is responsible for activating or deactivating the set of features $F$. Finally, the meta-heuristic intrinsic mechanics\footnote{Note that some meta-heuristic might start their searching procedure with every possible initial features, while others might start with a randomly subset of the initial features.} are executed to identify the best solution so far and to update the candidates' position in the hypercomplex space.

\subsection{Datasets}
\label{ss.datasets}

Table~\ref{t.datasets} describes all the datasets utilized in this work. We selected $20$ datasets that diversify within the number of samples, classes, and features, suggesting a more strong validation under distinguished scenarios. The datasets were downloaded from LibSVM's project\footnote{https://www.csie.ntu.edu.tw/\~cjlin/libsvmtools/datasets} and Arizona State University's (ASU) repository\footnote{http://featureselection.asu.edu/datasets.php}, being already quantized for categorical features and processed for missing values\footnote{One can find their post-processed versions at: http://recogna.tech}. As we need a unique set to guide the optimization process, not being the test one, we partitioned all datasets' training sets in half, composing the so-called validation set. Therefore, we use $25\%$ for the training step, $25\%$ for the optimization task validation, and the remaining $50\%$ to assess the experimental validation (testing step).

\begin{table}
\begin{center}
\renewcommand{\arraystretch}{1.5}
\setlength{\tabcolsep}{6pt}
\resizebox{350pt}{!}{
\begin{tabular}{lcccc}
\hhline{-|-|-|-|-}
\hhline{-|-|-|-|-}
\hhline{-|-|-|-|-}
\cellcolor[HTML]{EFEFEF}{\textbf{Dataset}}
& \cellcolor[HTML]{EFEFEF}{\textbf{Task}} & \cellcolor[HTML]{EFEFEF}{\textbf{\# Training Samples}} & \cellcolor[HTML]{EFEFEF}{\textbf{\# Testing Samples}} & \cellcolor[HTML]{EFEFEF}{\textbf{\# Features}}     \\ \hline
\multicolumn{1}{l}{Arcene}  & Mass Spectrometry & 100 &  100 &  10,000 \\ \hline
\multicolumn{1}{l}{BASEHOCK}  & Text &  997 &  996 &  4,862 \\ \hline
\multicolumn{1}{l}{COIL20}  & Face Image & 770  &  770 & 1,024 \\ \hline
\multicolumn{1}{l}{DNA} & Biological &  2,000 &  1,186 &  180 \\ \hline                 
\multicolumn{1}{l}{Isolet} & Spoken Letter Recognition &  780 &  780 &  617 \\ \hline
\multicolumn{1}{l}{Lung} & Biological &  102 &  101 &  3,312 \\ \hline
\multicolumn{1}{l}{Madelon} & Artificial & 2,000 &  600 &  500 \\ \hline
\multicolumn{1}{l}{MPEG7-BAS}  & Image Descriptor &  700 &  700 &  180 \\ \hline
\multicolumn{1}{l}{MPEG7-Fourier}  & Image Descriptor &  700 &  700 &  126 \\ \hline
\multicolumn{1}{l}{Mushrooms} & Biological &  4,062 &  4,062 &  112 \\ \hline
\multicolumn{1}{l}{NTL-Commercial} & Energy Theft &  2,476 &  2,476 &  8 \\ \hline
\multicolumn{1}{l}{NTL-Industrial} & Energy Theft &  1,591 &  1,591 &  8 \\ \hline
\multicolumn{1}{l}{ORL}  & Face Image &  200 &  200 &  1,024 \\ \hline
\multicolumn{1}{l}{PCMAC}  & Text &  972 &  971 &  3,289 \\ \hline
\multicolumn{1}{l}{Phishing} & Network Security &  5,528 &  5,527&  68 \\ \hline
\multicolumn{1}{l}{Segment}  & Image Segmentation &  1,155 &  1,155&  19 \\ \hline
\multicolumn{1}{l}{Sonar} & Signal &  104 &  104 & 60 \\ \hline
\multicolumn{1}{l}{Splice} & Biological &  1,000 &  2,175 &  60 \\ \hline
\multicolumn{1}{l}{Vehicle} & Image Silhouettes &  423 &  423 & 18 \\ \hline
\multicolumn{1}{l}{Wine} & Chemical &  89 &  89 & 13 \\
\hhline{-|-|-|-|-}
\hhline{-|-|-|-|-}
\hhline{-|-|-|-|-}
\end{tabular}}
\caption{Employed datasets used in the computations}.
\label{t.datasets}
\end{center}
\end{table}

\subsection{Computational Setup}
\label{ss.setup}

The source code used in this work comes from two libraries: LibOPT\footnote{https://github.com/jppbsi/LibOPT} and LibDEV\footnote{https://github.com/jppbsi/LibDEV}. Both libraries are implemented in the C language and have been extensively used throughout scientific research. The LibOPT library is a collection of meta-heuristic optimization techniques, while the LibDEV library provides an integration environment, e.g., feature selection conducted over meta-heuristic optimizations. One can refer to~\cite{Papa:18} in order to understand how it is possible to work under the LibOPT environment, i.e., how to design a hypercomplex optimization task.

To perform a reasonable comparison among distinct meta-heuristic techniques, we must rely on mathematical methods that will sustain these observations. The first step is to decide whether to use a parametric or a non-parametric statistical test~\cite{Hollander:13}. Unfortunately, we can not consider a normality state from our numerical trials due to the aleatory and non-deterministic factor derived from the meta-heuristic techniques, restraining our analysis to non-parametric approaches.

Secondly, acknowledging that the results of our numerical trials are independent (i.e., classification accuracy) and continuous over a particular dependent variable (i.e., number of observations), we can identify that the Wilcoxon signed-rank test~\cite{Wilcoxon:45} will satisfy our obligations. It is a non-parametric hypothesis test used to compare two or more related observations (in our case, repeated measurements over a certain meta-heuristic) to assess whether there are statistically significant differences between them.

\begin{sloppypar}
For every dataset, each meta-heuristic was evaluated under a 2-fold cross-validation\footnote{Remember that the training set was again split in half to form the validation set, used during the optimization process.} with 25 runs. Additionally, for every meta-heuristic, 15 agents (particles) were used over 25 convergence iterations. To provide a thorough comparison between meta-heuristics, we have chosen different techniques, ranging from swarm-based to evolutionary-inspired ones, in the context of feature selection:
\end{sloppypar}

\begin{itemize}
    \item Artificial Bee Colony (ABC)~\cite{Karaboga:07};
    \item Adaptive Inertia Weight Particle Swarm Optimization (AIWPSO)~\cite{NickabadiASC:11};
    \item Bat Algorithm (BA)~\cite{YangBA:12};
    \item Cuckoo Search (CS)~\cite{YangIJMMNO:10};
    \item Firefly Algorithm (FA)~\cite{YangFFA:10};
    \item Flower Pollination Algorithm (FPA)~\cite{YangFPA:14};
    \item Particle Swarm Optimization (PSO)~\cite{Kennedy:01}.
\end{itemize}

Note that, for each selected meta-heuristic, we will also present their quaternion- and octonion-based versions, being the former preceded by a Q prefix and the latter preceded by an O prefix. Table~\ref{t.parameters} presents the chosen parameter setting for every meta-heuristic technique\footnote{Note that these values were empirically chosen according to their authors' definition.}. We overlooked quaternion- and octonion-based algorithms from the table, as their parameters are the same as their original version.

\begin{table}
\begin{center}
\renewcommand{\arraystretch}{1.5}
\setlength{\tabcolsep}{6pt}
\resizebox{200pt}{!}{
\begin{tabular}{lr}
\hhline{-|-|}
\hhline{-|-|}
\hhline{-|-|}
\cellcolor[HTML]{EFEFEF}{\textbf{Algorithm}}
& \cellcolor[HTML]{EFEFEF}{\textbf{Parameters}}
\\ \hline
ABC & $\text{number of trials} = 1,000$\\ \hline
AIWPSO & $c_1 = 1.7 \mid c_2 = 1.7 \mid w = [0.5, 1.5]$\\ \hline
BA & $f = [0, 100] \mid A = 1.5 \mid r = 0.5$\\ \hline
CS & $\beta = 1.5 \mid p = 0.25 \mid \alpha = 0.8$\\ \hline
FA & $\alpha = 0.2 \mid \beta = 1.0 \mid \gamma = 1.0$\\ \hline
FPA & $\beta = 1.5 \mid p = 0.8$\\ \hline
PSO & $c_1 = 1.7 \mid c_2 = 1.7 \mid w = 0.7$\\
\hhline{-|-|}
\hhline{-|-|}
\hhline{-|-|}
\end{tabular}}
\end{center}
\caption{Parameter settings for the meta-heuristic algorithms considered in the work.}
\label{t.parameters}
\end{table}

Concerning ABC, we only need to set the number of trial limits for each food source. AIWPSO defines minimum and maximum weight as a $w$ interval, and $c_1$ and $c_2$ as the control parameters. BA has the minimum and maximum frequency ranges defined by $f$ interval, as well as the loudness parameter $A$ and pulse rate $r$. With CS, we demand to set up $\beta$, which is used to compute the L\'evy distribution, as well as $p$, which is the probability of replacing worst nests by new ones and $\alpha$ which is the step size. Regarding FA, we have $\alpha$ for calculating the randomized parameter, as well as the attractiveness parameter $\beta_0$ and the light absorption coefficient $\gamma$. FPA requires the $\beta$ parameter, used to compute the L\'evy distribution and $p$, which is the probability of local pollination. Finally, PSO defines $w$ as the inertia weight, and $c_1$ and $c_2$ as the control parameters.

%% file: results.tex
\section{Numerical Results}
\label{s.results}

This section presents the numerical results concerning the proposed experiments. Furthermore, it is divided into two subsections, which are in charge of discussing the overall analysis and the convergence analysis, respectively.

In order to provide statistical analysis to the numerical results, we opted to bold the best results' cells according to the Wilcoxon signed-rank test with 5\% of significance. In other words, it is possible to observe that, regarding a particular column, every bolded cell achieved the most suitable accuracy, time, or number of features according to the statistical test.

\subsection{Overall Analysis}
\label{ss.overall_analysis}

Table~\ref{t.accuracy} describes all datasets' average accuracy over the test set found by each meta-heuristic technique. A very interesting fact to highlight is that for almost every dataset, at least one meta-heuristic technique was able to achieve a performance comparable to the baseline approach, i.e., OPF classification using the whole dataset. On the other hand, considering Arcene, Mushrooms, NTL-Commercial, NTL-Industrial, and Splice datasets, meta-heuristic techniques outperformed the baseline approach.

Regarding only meta-heuristic techniques and their hypercomplex versions, one can see that the hypercomplex-based algorithms were able to achieve comparable accuracy values. In some cases, they even outperformed their na\"ive versions, e.g., QFA, QFPA, OFPA, and QPSO on Arcene; QAIWPSO on COIL20; QAIWPSO on Madelon; QCS, QFPA, and OFPA in Mushrooms; QABC and OABC in ORL; QABC, QBA, and OBA in PCMAC; QABC, OABC, OAIWPSO, QFA, OFA, QFPA and OFPA in Phishing; OABC and QPSO in Splice; QCS and OCS in Wine. In such case, outperforming means that a particular technique was capable of achieving a higher accuracy than another technique. Essentially, the Wilcoxon signed-rank test assess whether there was a statistical similarity between the accuracies obtained by each one of the techniques. Thus, as the statistical test was conducted over the independent accuracies for each meta-heuristic technique, it is possible to observe that the most significant techniques (bolded ones) in the aforementioned cases were also the ones that achieved higher accuracy than their na\"ive versions.

\begin{table}[!ht]
\begin{adjustwidth}{-2cm}{0cm}
\begin{center}
\renewcommand{\arraystretch}{1.75}
\setlength{\tabcolsep}{6pt}
\resizebox{575pt}{!}{
\begin{tabular}{lcccccccccccccccccccccc}
\hhline{-|-|-|-|-|-|-|-|-|-|-|-|-|-|-|-|-|-|-|-|-|-|-|}
\hhline{-|-|-|-|-|-|-|-|-|-|-|-|-|-|-|-|-|-|-|-|-|-|-|}
\hhline{-|-|-|-|-|-|-|-|-|-|-|-|-|-|-|-|-|-|-|-|-|-|-|}
\cellcolor[HTML]{EFEFEF}{} &
\cellcolor[HTML]{EFEFEF}{\textbf{ABC}} &
\cellcolor[HTML]{EFEFEF}{\textbf{QABC}} &
\cellcolor[HTML]{EFEFEF}{\textbf{OABC}} &
\cellcolor[HTML]{EFEFEF}{\textbf{AIWPSO}} &
\cellcolor[HTML]{EFEFEF}{\textbf{QAIWPSO}} &
\cellcolor[HTML]{EFEFEF}{\textbf{OAIWPSO}} &
\cellcolor[HTML]{EFEFEF}{\textbf{BA}} &
\cellcolor[HTML]{EFEFEF}{\textbf{QBA}} &
\cellcolor[HTML]{EFEFEF}{\textbf{OBA}} &
\cellcolor[HTML]{EFEFEF}{\textbf{CS}} &
\cellcolor[HTML]{EFEFEF}{\textbf{QCS}} &
\cellcolor[HTML]{EFEFEF}{\textbf{OCS}} &
\cellcolor[HTML]{EFEFEF}{\textbf{FA}} &
\cellcolor[HTML]{EFEFEF}{\textbf{QFA}} &
\cellcolor[HTML]{EFEFEF}{\textbf{OFA}} &
\cellcolor[HTML]{EFEFEF}{\textbf{FPA}} &
\cellcolor[HTML]{EFEFEF}{\textbf{QFPA}} &
\cellcolor[HTML]{EFEFEF}{\textbf{OFPA}} &
\cellcolor[HTML]{EFEFEF}{\textbf{PSO}} &
\cellcolor[HTML]{EFEFEF}{\textbf{QPSO}} &
\cellcolor[HTML]{EFEFEF}{\textbf{OPSO}} &
\cellcolor[HTML]{EFEFEF}{\textbf{BASELINE}}
\\ \hline
Arcene & \textbf{83.13\%} & \textbf{83.04\%} & \textbf{83.13\%} & \textbf{83.50\%} & 82.89\% & 82.52\% & 82.83\% & 82.72\% & 82.61\% & 82.41\% & 82.38\% & 82.75\% & 82.70\% & \textbf{83.04\%} & 82.55\% & 82.72\% & \textbf{82.90\%} & \textbf{82.92\%} & 82.26\% & \textbf{83.21\%} & 82.47\% & 81.58\%
\\ \hline
BASEHOCK & 79.49\% & 79.07\% & 79.98\% & 79.67\% & 79.86\% & 78.84\% & 79.99\% & 79.58\% & 79.73\% & 80.14\% & 78.87\% & 79.83\% & 79.25\% & 79.99\% & 79.62\% & 79.80\% & 79.00\% & 79.02\% & 79.63\% & 79.59\% & 80.13\% & \textbf{82.12\%}
\\ \hline
COIL20 & 99.04\% & 99.09\% & 99.03\% & 99.02\% & \textbf{99.13\%} & 99.07\% & 99.09\% & 99.08\% & 99.08\% & \textbf{99.12\%} & 99.09\% & 99.07\% & 99.08\% & 99.06\% & 99.09\% & 99.09\% & 99.04\% & 99.10\% & \textbf{99.13\%} & 99.07\% & 99.04\% & \textbf{99.30\%}
\\ \hline
DNA & 79.43\% & 79.58\% & 79.17\% & 78.95\% & 78.70\% & 78.64\% & 79.08\% & 79.32\% & 77.55\% & 77.90\% & 77.49\% & 78.06\% & 78.69\% & 79.23\% & 78.97\% & 79.17\% & 79.30\% & 79.30\% & 79.51\% & 78.71\% & 79.26\% & \textbf{82.33\%}
\\ \hline
Isolet & 90.79\% & 90.81\% & 90.72\% & 90.89\% & 90.77\% & 90.83\% & 90.55\% & 90.71\% & 90.77\% & 90.57\% & 90.71\% & 90.70\% & 90.68\% & 90.89\% & 90.73\% & 90.77\% & 90.67\% & 90.69\% & 90.67\% & 90.71\% & 90.66\% & \textbf{91.30\%}
\\ \hline
Lung & \textbf{93.08\%} & \textbf{93.13\%} & \textbf{92.68\%} & \textbf{91.95\%} & \textbf{92.43\%} & \textbf{92.26\%} & \textbf{92.61\%} & \textbf{92.66\%} & \textbf{92.77\%} & \textbf{92.20\%} & \textbf{92.47\%} & \textbf{92.56\%} & \textbf{92.70\%} & \textbf{92.54\%} & \textbf{92.49\%} & \textbf{92.88\%} & \textbf{92.71\%} & \textbf{92.87\%} & \textbf{92.83\%} & \textbf{91.89\%} & \textbf{92.51\%} & \textbf{93.24\%}
\\ \hline
Madelon & \textbf{63.42\%} & 62.69\% & 63.00\% & 62.73\% & \textbf{64.52\%} & 63.21\% & \textbf{63.85\%} & 62.25\% & \textbf{65.12\%} & \textbf{63.48\%} & 62.69\% & \textbf{63.63\%} & \textbf{63.94\%} & 63.23\% & 62.79\% & \textbf{62.81\%} & \textbf{63.70\%} & 63.41\% & 62.87\% & 63.03\% & 62.47\% & \textbf{64.37\%}
\\ \hline
MPEG7-BAS & 88.78\% & \textbf{88.85\%} & \textbf{88.90\%} & \textbf{88.91\%} & \textbf{88.90\%} & \textbf{88.90\%} & \textbf{88.80\%} & \textbf{88.98\%} & \textbf{88.81\%} & \textbf{88.92\%} & \textbf{88.97\%} & \textbf{88.81\%} & \textbf{88.89\%} & \textbf{88.99\%} & \textbf{88.83\%} & \textbf{88.81\%} & \textbf{88.85\%} & 88.81\% & \textbf{88.90\%} & \textbf{88.96\%} & \textbf{88.88\%} & \textbf{89.11\%}
\\ \hline
MPEG7-Fourier & \textbf{72.12\%} & \textbf{71.94\%} & \textbf{72.00\%} & \textbf{71.91\%} & \textbf{71.91\%} & \textbf{72.18\%} & \textbf{72.19\%} & \textbf{72.06\%} & \textbf{72.18\%} & 69.46\% & 70.00\% & 70.49\% & 71.78\% & 71.86\% & \textbf{71.96\%} & \textbf{72.02\%} & \textbf{72.19\%} & \textbf{72.16\%} & \textbf{72.12\%} & \textbf{72.22\%} & \textbf{72.14\%} & \textbf{72.09\%}
\\ \hline
Mushrooms & \textbf{96.19\%} & \textbf{96.57\%} & \textbf{98.81\%} & \textbf{96.61\%} & \textbf{97.63\%} & \textbf{97.43\%} & \textbf{95.81\%} & \textbf{96.61\%} & 94.04\% & 97.11\% & \textbf{97.40\%} & 94.62\% & \textbf{96.90\%} & \textbf{94.80\%} & \textbf{96.31\%} & 94.14\% & \textbf{95.68\%} & \textbf{97.09\%} & \textbf{96.09\%} & 95.39\% & \textbf{96.09\%} & 94.36\%
\\ \hline
NTL-Commercial & \textbf{92.73\%} & \textbf{92.59\%} & \textbf{92.16\%} & \textbf{91.84\%} & 90.78\% & \textbf{92.39\%} & \textbf{92.42\%} & \textbf{92.17\%} & \textbf{91.93\%} & 79.22\% & 76.07\% & 79.99\% & \textbf{91.71\%} & \textbf{91.99\%} & \textbf{92.05\%} & \textbf{92.17\%} & \textbf{92.39\%} & \textbf{91.76\%} & \textbf{92.52\%} & \textbf{92.57\%} & \textbf{91.93\%} & 61.45\%
\\ \hline
NTL-Industrial & \textbf{95.77\%} & \textbf{95.36\%} & \textbf{95.31\%} & \textbf{95.40\%} & \textbf{95.37\%} & \textbf{95.42\%} & \textbf{95.58\%} & \textbf{95.04\%} & \textbf{95.22\%} & 77.90\% & 82.95\% & 79.16\% & \textbf{94.80\%} & \textbf{95.82\%} & \textbf{94.64\%} & \textbf{95.05\%} & 94.79\% & \textbf{94.77\%} & \textbf{94.99\%} & \textbf{94.84\%} & \textbf{94.54\%} & 67.86\%
\\ \hline
ORL & 93.56\% & \textbf{93.75\%} & \textbf{93.60\%} & \textbf{93.67\%} & 93.54\% & 93.46\% & \textbf{93.57\%} & \textbf{93.84\%} & \textbf{93.57\%} & \textbf{93.62\%} & \textbf{93.60\%} & \textbf{93.53\%} & \textbf{93.58\%} & 93.60\% & \textbf{93.69\%} & \textbf{93.66\%} & 93.56\% & \textbf{93.72\%} & \textbf{93.75\%} & 93.57\% & 93.49\% & \textbf{93.25\%}
\\ \hline
PCMAC & 71.97\% & \textbf{72.24\%} & 71.35\% & \textbf{72.44\%} & 71.16\% & 71.58\% & 71.18\% & \textbf{71.83\%} & \textbf{73.18\%} & \textbf{72.03\%} & \textbf{72.32\%} & 71.98\% & \textbf{71.99\%} & 71.50\% & \textbf{72.19\%} & \textbf{72.78\%} & \textbf{72.54\%} & \textbf{72.36\%} & \textbf{71.81\%} & 71.64\% & \textbf{72.35\%} & \textbf{72.75\%}
\\ \hline
Phishing & 84.53\% & \textbf{85.36\%} & \textbf{86.56\%} & 84.74\% & 84.02\% & \textbf{85.99\%} & \textbf{84.92\%} & 85.49\% & \textbf{84.76\%} & 85.40\% & 84.90\% & 84.06\% & 83.35\% & \textbf{85.26\%} & \textbf{85.19\%} & 84.40\% & \textbf{84.40\%} & \textbf{85.61\%} & \textbf{84.79\%} & 84.10\% & \textbf{85.93\%} & \textbf{86.67\%}
\\ \hline
Segment & \textbf{97.33\%} & \textbf{97.19\%} & \textbf{97.25\%} & \textbf{97.15\%} & \textbf{97.05\%} & \textbf{97.20\%} & \textbf{97.19\%} & \textbf{97.20\%} & \textbf{97.06\%} & 96.22\% & 96.35\% & 95.95\% & \textbf{97.29\%} & 96.82\% & \textbf{97.21\%} & \textbf{97.13\%} & \textbf{97.20\%} & \textbf{97.19\%} & \textbf{97.17\%} & \textbf{97.31\%} & 96.83\% & \textbf{97.34\%}
\\ \hline
Sonar & \textbf{79.40\%} & \textbf{81.72\%} & \textbf{79.57\%} & \textbf{81.21\%} & \textbf{81.56\%} & \textbf{79.89\%} & \textbf{81.20\%} & \textbf{80.15\%} & \textbf{80.62\%} & \textbf{80.44\%} & \textbf{80.94\%} & \textbf{80.76\%} & \textbf{80.32\%} & \textbf{80.20\%} & \textbf{80.76\%} & \textbf{79.80\%} & \textbf{80.53\%} & \textbf{80.80\%} & \textbf{80.78\%} & \textbf{80.14\%} & \textbf{80.77\%} & \textbf{81.75\%}
\\ \hline
Splice & 71.64\% & 71.33\% & \textbf{71.99\%} & \textbf{71.87\%} & \textbf{71.72\%} & 70.99\% & 70.93\% & 71.14\% & 71.32\% & 69.53\% & 69.56\% & 70.21\% & \textbf{71.32\%} & 71.33\% & \textbf{71.50\%} & \textbf{72.30\%} & \textbf{71.83\%} & 71.08\% & 70.81\% & \textbf{71.68\%} & 71.35\% & 70.89\%
\\ \hline
Vehicle & \textbf{78.17\%} & \textbf{77.41\%} & \textbf{77.62\%} & \textbf{77.11\%} & 77.25\% & \textbf{77.42\%} & \textbf{77.60\%} & \textbf{77.51\%} & 76.98\% & 76.74\% & 76.20\% & 76.34\% & \textbf{77.25\%} & 76.79\% & \textbf{77.11\%} & \textbf{77.46\%} & \textbf{77.09\%} & \textbf{77.71\%} & 76.93\% & 76.53\% & 76.85\% & \textbf{77.63\%}
\\ \hline
Wine & \textbf{96.21\%} & \textbf{96.46\%} & \textbf{96.58\%} & \textbf{95.94\%} & 95.62\% & \textbf{95.96\%} & \textbf{96.67\%} & \textbf{96.63\%} & \textbf{96.23\%} & 94.87\% & \textbf{95.51\%} & \textbf{95.55\%} & \textbf{96.42\%} & \textbf{96.22\%} & \textbf{95.99\%} & \textbf{96.16\%} & \textbf{96.09\%} & 95.81\% & \textbf{96.23\%} & \textbf{96.39\%} & \textbf{95.98\%} & \textbf{95.80\%}
\\
\hhline{-|-|-|-|-|-|-|-|-|-|-|-|-|-|-|-|-|-|-|-|-|-|-|}
\hhline{-|-|-|-|-|-|-|-|-|-|-|-|-|-|-|-|-|-|-|-|-|-|-|}
\hhline{-|-|-|-|-|-|-|-|-|-|-|-|-|-|-|-|-|-|-|-|-|-|-|}
\end{tabular}}
\end{center}
\caption{Average accuracy achieved over the test set considering all datasets.}
\label{t.accuracy}
\end{adjustwidth}
\end{table}

An interesting fact emerging from Table~\ref{t.features} is that CS was able to achieve the lowest number of features in nearly every dataset. Furthermore, when standard CS did not deliver the lowest number of features, its quaternionic and octatonic representations were able to achieve this intent\footnote{QCS achieved the lowest number of features for the Phishing dataset while OCS achieved this goal for the DNA, NTL-Commercial and Segment datasets.}.

Even though most algorithms were able to diminish the features' space size and obtain statistically similar accuracy within respect to the baseline method, in some cases they reached a slightly lower accuracy than the original OPF classification. However, it should be noted that in the case of the BASEHOCK dataset, where even the baseline classification obtained the best accuracy, all other meta-heuristic techniques could reduce by about 35\% of the number of features while scoring 2-3\% lower accuracy than OPF.

\begin{table}[!ht]
\begin{adjustwidth}{-2cm}{0cm}
\begin{center}
\renewcommand{\arraystretch}{1.75}
\setlength{\tabcolsep}{6pt}
\resizebox{575pt}{!}{
\begin{tabular}{lcccccccccccccccccccccc}
\hhline{-|-|-|-|-|-|-|-|-|-|-|-|-|-|-|-|-|-|-|-|-|-|-|}
\hhline{-|-|-|-|-|-|-|-|-|-|-|-|-|-|-|-|-|-|-|-|-|-|-|}
\hhline{-|-|-|-|-|-|-|-|-|-|-|-|-|-|-|-|-|-|-|-|-|-|-|}
\cellcolor[HTML]{EFEFEF}{} &
\cellcolor[HTML]{EFEFEF}{\textbf{ABC}} &
\cellcolor[HTML]{EFEFEF}{\textbf{QABC}} &
\cellcolor[HTML]{EFEFEF}{\textbf{OABC}} &
\cellcolor[HTML]{EFEFEF}{\textbf{AIWPSO}} &
\cellcolor[HTML]{EFEFEF}{\textbf{QAIWPSO}} &
\cellcolor[HTML]{EFEFEF}{\textbf{OAIWPSO}} &
\cellcolor[HTML]{EFEFEF}{\textbf{BA}} &
\cellcolor[HTML]{EFEFEF}{\textbf{QBA}} &
\cellcolor[HTML]{EFEFEF}{\textbf{OBA}} &
\cellcolor[HTML]{EFEFEF}{\textbf{CS}} &
\cellcolor[HTML]{EFEFEF}{\textbf{QCS}} &
\cellcolor[HTML]{EFEFEF}{\textbf{OCS}} &
\cellcolor[HTML]{EFEFEF}{\textbf{FA}} &
\cellcolor[HTML]{EFEFEF}{\textbf{QFA}} &
\cellcolor[HTML]{EFEFEF}{\textbf{OFA}} &
\cellcolor[HTML]{EFEFEF}{\textbf{FPA}} &
\cellcolor[HTML]{EFEFEF}{\textbf{QFPA}} &
\cellcolor[HTML]{EFEFEF}{\textbf{OFPA}} &
\cellcolor[HTML]{EFEFEF}{\textbf{PSO}} &
\cellcolor[HTML]{EFEFEF}{\textbf{QPSO}} &
\cellcolor[HTML]{EFEFEF}{\textbf{OPSO}} &
\cellcolor[HTML]{EFEFEF}{\textbf{BASELINE}}
\\ \hline
Arcene & 6462.88 & 6473.48 & 6482.36 & 6454.20 & 6492.80 & 6467.04 & 6476.72 & 6486.52 & 6512.96 & \textbf{6419.40} & 6453.76 & 6478.40 & 6466.00 & 6471.68 & 6491.80 & 6471.56 & 6481.52 & 6479.60 & 6450.84 & 6468.56 & 6505.52 & 10000.00
\\ \hline
BASEHOCK & 3166.92 & 3141.28 & 3142.16 & 3148.48 & 3149.76 & 3153.28 & 3167.52 & 3167.60 & 3170.32 & \textbf{3124.72} & 3143.72 & 3139.64 & 3152.16 & 3146.40 & 3156.52 & 3151.52 & 3131.08 & 3135.00 & 3149.40 & 3147.32 & 3147.68 & 4862.00
\\ \hline
COIL20 & 670.04 & 666.16 & 663.92 & 658.92 & 664.12 & 668.84 & 667.56 & 668.28 & 674.88 & \textbf{654.52} & 660.52 & 664.56 & 666.12 & 664.44 & 667.00 & 663.56 & 664.08 & 666.44 & 664.40 & 666.80 & 666.88 & 1024.00
\\ \hline
DNA & 117.04 & 117.48 & 116.44 & 117.20 & 116.80 & 117.88 & 119.96 & 116.92 & 116.72 & 118.04 & 116.36 & \textbf{113.00} & 117.84 & 116.40 & 115.96 & 119.04 & 117.96 & 119.00 & 115.00 & 116.28 & 117.00 & 180.00
\\ \hline
Isolet & 407.08 & 401.12 & 398.16 & 396.24 & 399.12 & 401.08 & 399.48 & 403.36 & 408.84 & \textbf{393.40} & 398.64 & 401.40 & 402.52 & 397.88 & 399.76 & 402.20 & 397.68 & 400.32 & 397.24 & 397.88 & 399.72 & 617.00
\\ \hline
Lung & 2138.28 & 2132.52 & 2135.64 & \textbf{2125.32} & 2143.64 & 2147.60 & 2148.08 & 2151.12 & 2148.36 & 2130.36 & 2144.88 & 2143.84 & 2150.40 & 2143.88 & 2143.64 & 2136.52 & 2141.64 & 2147.96 & 2134.40 & 2145.24 & 2141.92 & 3312.00
\\ \hline
Madelon & 324.72 & \textbf{321.72} & 324.40 & 321.96 & 322.44 & 324.44 & 324.00 & 327.44 & 330.36 & 325.16 & 324.24 & 324.44 & 325.84 & 328.80 & 323.16 & 324.68 & 323.48 & 327.56 & 324.64 & 323.92 & 324.00 & 500.00
\\ \hline
MPEG7-BAS & 118.12 & 120.80 & 119.00 & 118.40 & 120.12 & 119.84 & 118.56 & 121.52 & 120.44 & \textbf{115.88} & 116.84 & 116.80 & 120.92 & 117.68 & 119.20 & 118.32 & 118.92 & 117.88 & 117.88 & 118.52 & 119.44 & 180.00
\\ \hline
MPEG7-Fourier & 81.08 & 82.08 & 81.88 & 82.52 & \textbf{81.04} & 82.36 & 84.36 & 84.40 & 83.08 & 81.12 & 81.48 & 82.64 & 82.64 & 82.92 & 81.24 & 82.36 & 82.84 & 83.04 & 83.88 & 82.44 & 81.32 & 126.00
\\ \hline
Mushrooms & 74.92 & 70.36 & 70.56 & 72.48 & 74.04 & 73.28 & 73.44 & 72.88 & 72.84 & \textbf{70.32} & 73.08 & 71.76 & 74.96 & 72.16 & 73.12 & 75.24 & 74.12 & 70.52 & 73.08 & 73.28 & 73.84 & 112.00
\\ \hline
NTL-Commercial & 5.72 & 5.60 & 5.60 & 5.84 & 5.76 & 5.76 & 5.68 & 5.76 & 5.56 & 5.04 & 5.04 & \textbf{4.84} & 5.68 & 5.72 & 5.44 & 5.64 & 5.68 & 5.64 & 5.68 & 5.48 & 5.44 & 8.00
\\ \hline
NTL-Industrial & 5.40 & 5.24 & 5.52 & 5.24 & 5.36 & 5.44 & 5.68 & 5.80 & 5.44 & 5.76 & 5.20 & 5.08 & 5.56 & \textbf{5.00} & 5.40 & 5.20 & 5.20 & 5.32 & 5.52 & 5.64 & 5.44 & 8.00
\\ \hline
ORL & 665.76 & 656.96 & 659.96 & 662.96 & 663.28 & 663.88 & 666.48 & 665.48 & 671.24 & \textbf{653.92} & 661.00 & 659.84 & 664.72 & 658.40 & 665.56 & 662.84 & 657.80 & 662.80 & 658.16 & 660.76 & 659.16 & 1024.00
\\ \hline
PCMAC & 2152.68 & 2114.28 & 2130.60 & 2132.16 & 2123.76 & 2128.32 & 2140.04 & 2137.88 & 2154.60 & \textbf{2112.16} & 2115.40 & 2123.12 & 2129.24 & 2126.48 & 2130.44 & 2135.56 & 2128.88 & 2130.92 & 2128.48 & 2126.76 & 2131.52 & 3289.00
\\ \hline
Phishing & 45.16 & 44.68 & 45.60 & 44.52 & 44.84 & 46.36 & 45.28 & 45.12 & 44.88 & 44.40 & \textbf{44.12} & 44.88 & 44.92 & 45.44 & 46.60 & 44.80 & 46.44 & 45.40 & 45.12 & 44.80 & 46.48 & 68.00
\\ \hline
Segment & 14.04 & 13.20 & 13.48 & 13.44 & 13.44 & 13.80 & 13.44 & 13.84 & 13.64 & 13.04 & 13.24 & \textbf{12.40} & 13.32 & 13.48 & 14.08 & 13.68 & 13.04 & 13.44 & 13.56 & 13.92 & 13.44 & 19.00
\\ \hline
Sonar & 40.24 & 39.40 & 38.80 & 39.00 & 39.76 & 40.84 & 40.04 & 37.96 & 39.12 & 39.52 & 39.64 & 39.48 & \textbf{37.92} & 38.24 & 40.28 & 39.76 & 39.08 & 40.32 & 39.48 & 38.24 & 39.96 & 60.00
\\ \hline
Splice & 40.48 & 39.72 & 39.72 & 40.68 & 40.24 & 40.28 & 40.56 & 38.80 & 40.64 & \textbf{38.76} & 39.40 & 40.04 & 39.28 & 40.40 & 40.92 & 40.40 & 39.92 & 40.48 & 40.04 & 40.44 & 39.80 & 60.00
\\ \hline
Vehicle & 13.40 & 12.68 & 12.72 & 12.84 & 12.00 & 12.28 & 12.60 & 12.96 & 13.40 & \textbf{11.92} & 12.20 & 12.08 & 12.84 & 12.72 & 13.00 & 13.08 & 13.28 & 12.76 & 13.52 & 12.52 & 13.16 & 18.00
\\ \hline
Wine & 9.20 & 9.04 & 9.24 & \textbf{8.40} & 9.04 & 9.12 & 9.28 & 9.00 & 9.56 & 9.04 & 8.80 & 8.88 & 9.04 & 9.08 & 9.48 & 9.00 & 9.32 & 8.76 & 9.28 & 8.92 & 9.48 & 13.00
\\
\hhline{-|-|-|-|-|-|-|-|-|-|-|-|-|-|-|-|-|-|-|-|-|-|-|}
\hhline{-|-|-|-|-|-|-|-|-|-|-|-|-|-|-|-|-|-|-|-|-|-|-|}
\hhline{-|-|-|-|-|-|-|-|-|-|-|-|-|-|-|-|-|-|-|-|-|-|-|}
\end{tabular}}
\end{center}
\caption{Average number of features used over the test set considering all datasets.}
\label{t.features}
\end{adjustwidth}
\end{table}

\begin{figure}[!ht]
	\centering
	\begin{adjustwidth}{-2cm}{0cm}
	\begin{tabular}{cc}
		\includegraphics[scale=0.25]{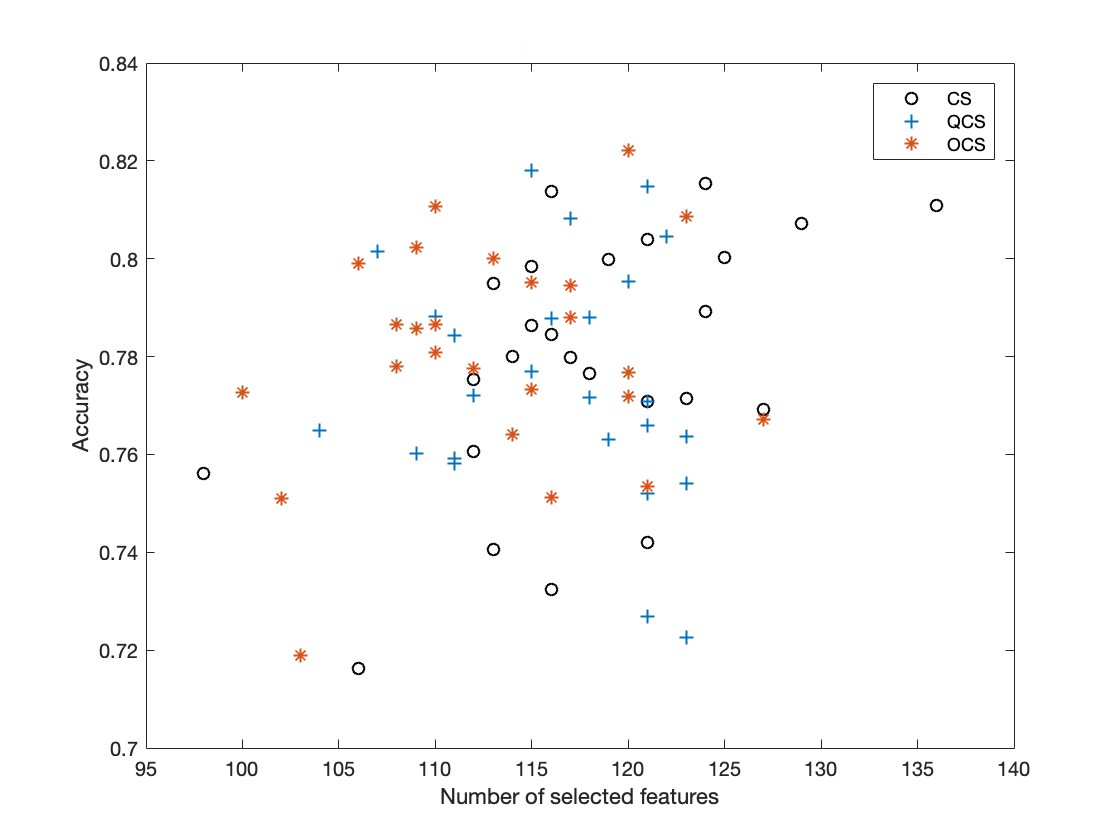} &
		\includegraphics[scale=0.25]{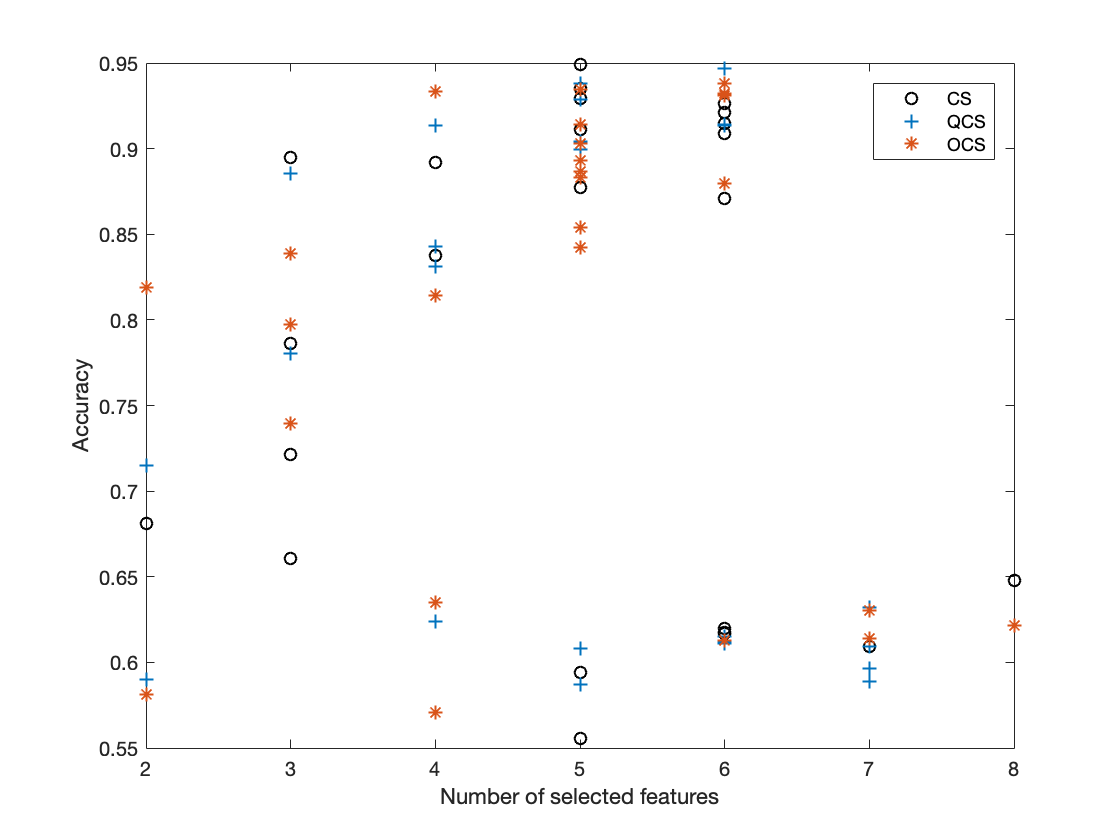}
		\\
		(a) & (b)
		\\
		\includegraphics[scale=0.25]{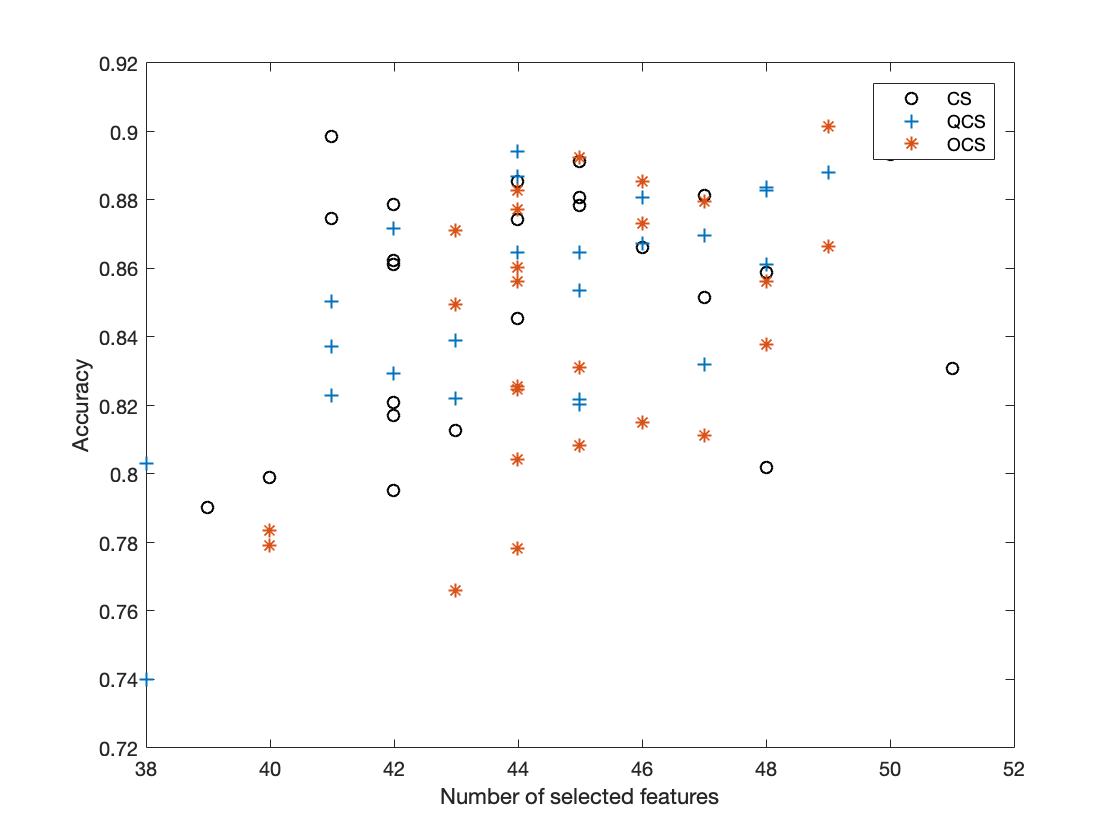} &
		\includegraphics[scale=0.25]{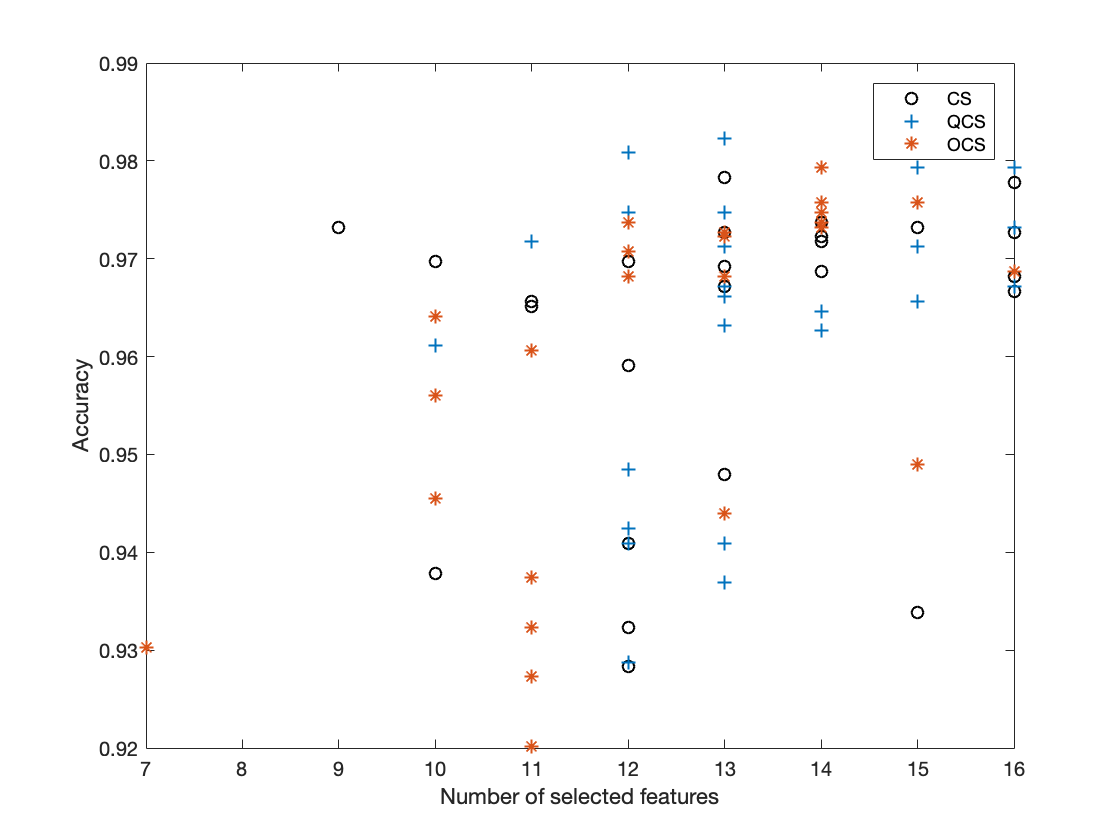}
		\\
		(c) & (d)
	\end{tabular}
	\end{adjustwidth}
	\caption{Number of selected features x Accuracy ([0,1]) chart considering CS, QCS and OCS in: (a) DNA, (b) NTL-Commercial, (c) Phishing and (d) Segment datasets.}
	\label{f.cs_acc_features}
 \end{figure}

\begin{figure}[!ht]
	\centering
	\begin{adjustwidth}{-2cm}{0cm}
	\begin{tabular}{cc}
		\includegraphics[scale=0.25]{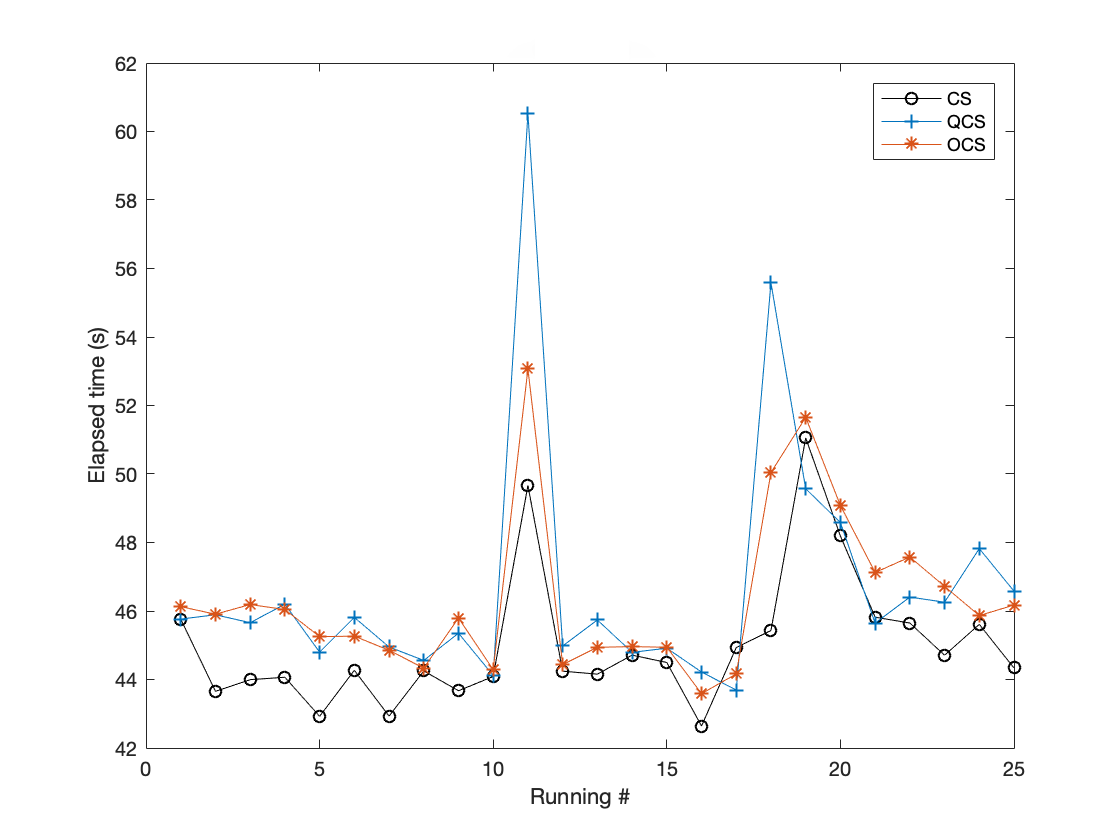} &
		\includegraphics[scale=0.25]{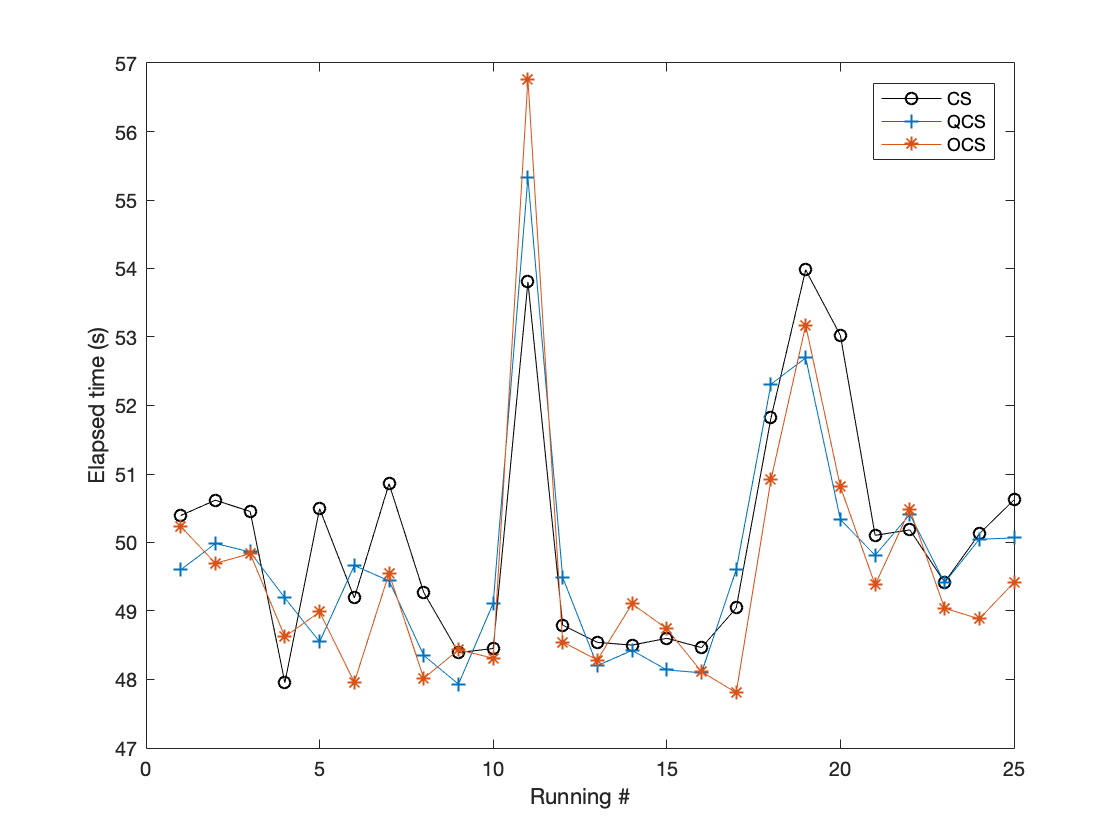}
		\\
		(a) & (b)
		\\
		\includegraphics[scale=0.25]{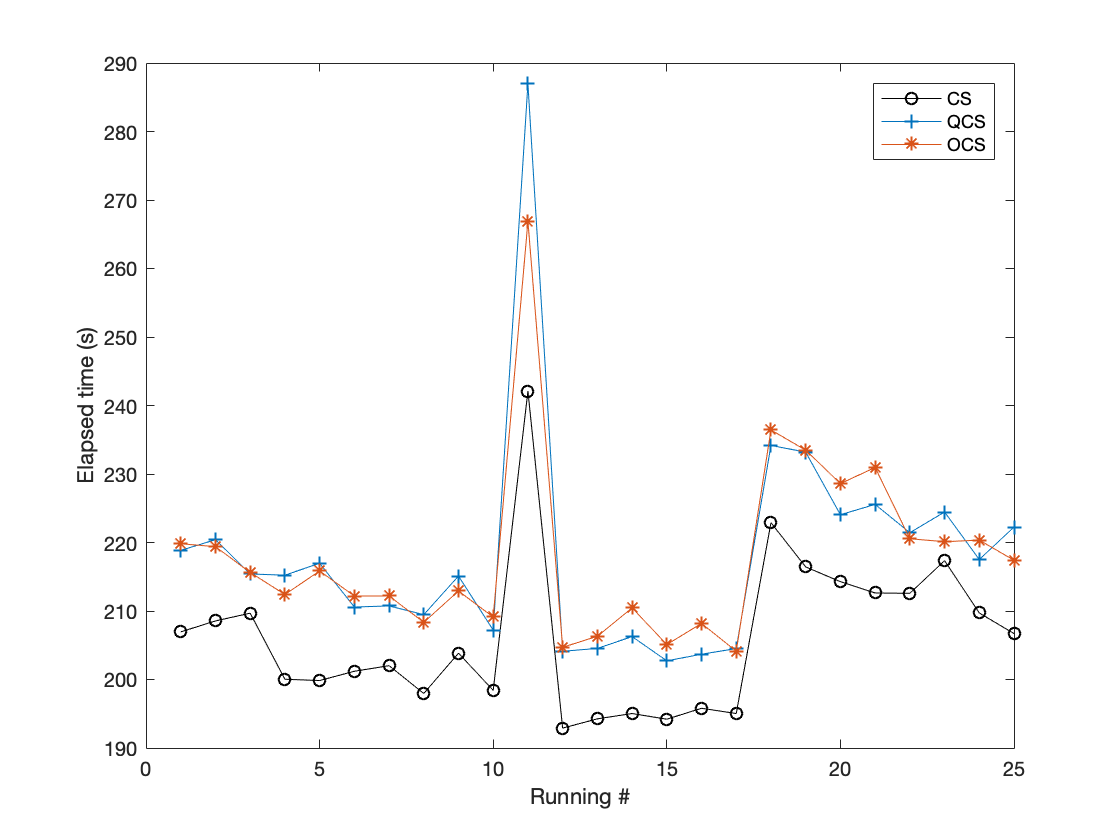} &
		\includegraphics[scale=0.25]{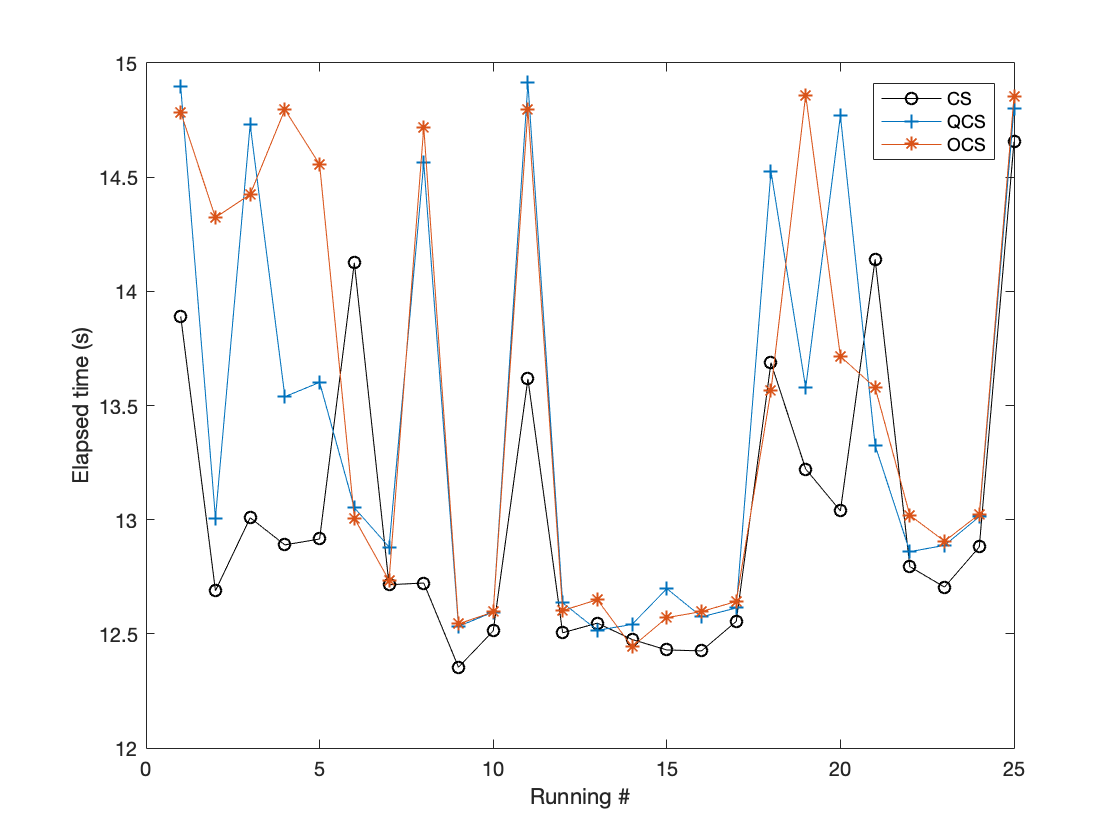}
		\\
		(c) & (d)
	\end{tabular}
	\end{adjustwidth}
	\caption{Computation time (s) for each independente run of CS, QCS and OCS in: (a) DNA, (b) NTL-Commercial, (c) Phishing and (d) Segment datasets.}
	\label{f.cs_time}
 \end{figure}

Table~\ref{t.time} shows that CS-based techniques completed the optimization runs in a significantly lower computation time than every other technique. Additionally, Figures~\ref{f.cs_acc_features} and~\ref{f.cs_time} illustrate a more in-depth comparison between CS and its hypercomplex versions for four distinct datasets: DNA, NTL-Commercial, Phishing and Segment. One can observe that Figure~\ref{f.cs_acc_features} represents the CS-based techniques behavior, where the best techniques are positioned in the top-left corner of the graphic, i.e., best accuracy and lowest number of features. For the sake of brevity, we opted to show some datasets that have discrepant data, i.e., datasets that have a low amount of features, being more susceptible when selecting a subset of features. In such cases, any incorrect feature selection will depreciate the classification results, thus, making the convergence process more unstable.

One can perceive that CS-based techniques encountered a feasible number of features, but not necessarily the best accuracy. If one observes the difference between the best and the worst accuracy considering all datasets (except NTL-based ones) and meta-heuristic techniques, there is not a single one that surpasses the 4.77\% barrier. Nevertheless, CS suffered in the NTL datasets (energy theft identification), which are highly unbalanced and have a relatively small amount of features. As CS encountered the lowest number of features in such a low-dimensional dataset, it is possible to observe that is has overfitted the optimization process to find the lowest number of possible features at the cost of penalizing the classifier, hence, achieving a not suitable accuracy for these particular datasets.

\begin{table}[!ht]
\begin{adjustwidth}{-2cm}{0cm}
\begin{center}
\renewcommand{\arraystretch}{1.75}
\setlength{\tabcolsep}{6pt}
\resizebox{575pt}{!}{
\begin{tabular}{lccccccccccccccccccccc}
\hhline{-|-|-|-|-|-|-|-|-|-|-|-|-|-|-|-|-|-|-|-|-|-|}
\hhline{-|-|-|-|-|-|-|-|-|-|-|-|-|-|-|-|-|-|-|-|-|-|}
\hhline{-|-|-|-|-|-|-|-|-|-|-|-|-|-|-|-|-|-|-|-|-|-|}
\cellcolor[HTML]{EFEFEF}{} &
\cellcolor[HTML]{EFEFEF}{\textbf{ABC}} &
\cellcolor[HTML]{EFEFEF}{\textbf{QABC}} &
\cellcolor[HTML]{EFEFEF}{\textbf{OABC}} &
\cellcolor[HTML]{EFEFEF}{\textbf{AIWPSO}} &
\cellcolor[HTML]{EFEFEF}{\textbf{QAIWPSO}} &
\cellcolor[HTML]{EFEFEF}{\textbf{OAIWPSO}} &
\cellcolor[HTML]{EFEFEF}{\textbf{BA}} &
\cellcolor[HTML]{EFEFEF}{\textbf{QBA}} &
\cellcolor[HTML]{EFEFEF}{\textbf{OBA}} &
\cellcolor[HTML]{EFEFEF}{\textbf{CS}} &
\cellcolor[HTML]{EFEFEF}{\textbf{QCS}} &
\cellcolor[HTML]{EFEFEF}{\textbf{OCS}} &
\cellcolor[HTML]{EFEFEF}{\textbf{FA}} &
\cellcolor[HTML]{EFEFEF}{\textbf{QFA}} &
\cellcolor[HTML]{EFEFEF}{\textbf{OFA}} &
\cellcolor[HTML]{EFEFEF}{\textbf{FPA}} &
\cellcolor[HTML]{EFEFEF}{\textbf{QFPA}} &
\cellcolor[HTML]{EFEFEF}{\textbf{OFPA}} &
\cellcolor[HTML]{EFEFEF}{\textbf{PSO}} &
\cellcolor[HTML]{EFEFEF}{\textbf{QPSO}} &
\cellcolor[HTML]{EFEFEF}{\textbf{OPSO}}
\\ \hline
Arcene & 43.48s & 42.11s & 43.06s & 21.59s & 23.08s & 23.45s & 21.92s & 23.97s & 25.78s & \textbf{7.81s} & 10.06s & 11.68s & 22.90s & 27.62s & 34.57s & 22.59s & 27.62s & 31.83s & 22.21s & 22.70s & 23.42s
\\ \hline
BASEHOCK & 1114.33s & 1105.60s & 1108.30s & 563.38s & 563.89s & 566.88s & 568.81s & 571.82s & 571.79s & \textbf{189.89s} & 203.89s & 203.53s & 545.73s & 543.84s & 547.47s & 564.81s & 565.72s & 567.54s & 556.54s & 558.63s & 561.71s
\\ \hline
COIL20 & 179.37s & 181.11s & 181.33s & 90.92s & 93.17s & 93.01s & 90.47s & 92.77s & 94.90s & \textbf{30.19s} & 33.10s & 33.57s & 89.41s & 88.91s & 90.34s & 92.62s & 92.79s & 94.25s & 90.70s & 91.86s & 91.87s
\\ \hline
DNA & 240.43s & 241.94s & 242.54s & 171.38s & 171.86s & 171.85s & 124.60s & 123.88s & 123.78s & \textbf{45.01s} & 46.74s & 46.34s & 121.66s & 122.14s & 122.58s & 123.84s & 125.02s & 124.48s & 171.20s & 171.06s & 169.50s
\\ \hline
Isolet & 135.52s & 138.87s & 138.42s & 69.96s & 70.40s & 70.60s & 69.60s & 71.11s & 71.01s & \textbf{21.85s} & 25.73s & 26.05s & 68.44s & 68.02s & 69.73s & 70.96s & 71.09s & 70.74s & 68.78s & 69.86s & 69.86s
\\ \hline
Lung & 15.19s & 15.07s & 15.50s & 7.84s & 8.12s & 8.42s & 7.73s & 8.60s & 9.01s & \textbf{2.76s} & 3.61s & 4.17s & 7.98s & 9.61s & 11.74s & 7.98s & 9.87s & 11.24s & 7.71s & 7.78s & 8.05s
\\ \hline
Madelon & 704.37s & 698.42s & 698.56s & 354.86s & 356.56s & 357.16s & 358.11s & 359.54s & 361.50s & \textbf{120.80s} & 127.19s & 128.23s & 343.79s & 341.46s & 343.17s & 355.31s & 353.68s & 354.26s & 351.40s & 352.89s & 352.85s
\\ \hline
MPEG7-BAS & 38.05s & 35.59s & 37.25s & 25.01s & 25.71s & 25.25s & 18.78s & 18.46s & 18.67s & \textbf{6.61s} & 7.29s & 7.39s & 18.96s & 19.36s & 19.17s & 19.51s & 18.89s & 19.64s & 25.40s & 25.16s & 25.06s
\\ \hline
MPEG7-Fourier & 27.92s & 25.70s & 25.78s & 19.29s & 19.54s & 19.46s & 13.74s & 13.87s & 13.97s & \textbf{4.68s} & 5.00s & 5.02s & 13.31s & 13.33s & 13.25s & 13.37s & 13.17s & 13.26s & 18.91s & 18.69s & 18.87s
\\ \hline
Mushrooms & 791.66s & 785.72s & 785.15s & 600.75s & 600.03s & 601.13s & 405.80s & 408.04s & 409.10s & \textbf{143.20s} & 151.30s & 152.22s & 398.83s & 397.73s & 398.85s & 404.05s & 403.06s & 404.90s & 597.34s & 596.48s & 595.58s
\\ \hline
NTL-Commercial & 272.72s & 271.35s & 270.80s & 139.44s & 138.38s & 138.36s & 139.15s & 138.73s & 138.41s & 50.04s & 49.76s & \textbf{49.56s} & 132.70s & 133.49s & 133.30s & 138.46s & 137.85s & 138.23s & 138.58s & 137.60s & 137.83s
\\ \hline
NTL-Industrial & 113.70s & 111.91s & 111.74s & 57.60s & 56.71s & 56.55s & 57.79s & 57.30s & 57.27s & 20.58s & 20.62s & \textbf{20.55s} & 54.88s & 54.97s & 54.58s & 57.11s & 57.30s & 57.02s & 57.68s & 56.60s & 56.46s
\\ \hline
ORL & 13.07s & 13.05s & 13.08s & 6.73s & 6.77s & 6.93s & 6.38s & 6.96s & 7.20s & \textbf{2.07s} & 2.62s & 2.82s & 6.67s & 7.56s & 8.63s & 6.87s & 7.19s & 7.67s & 6.64s & 6.79s & 6.80s
\\ \hline
PCMAC & 751.25s & 742.09s & 744.23s & 377.87s & 381.31s & 380.74s & 382.81s & 382.50s & 386.86s & \textbf{128.28s} & 136.87s & 137.31s & 365.66s & 366.69s & 370.00s & 380.84s & 382.02s & 381.16s & 374.06s & 376.45s & 377.26s
\\ \hline
Phishing & 1119.79s & 1110.41s & 1110.58s & 943.66s & 941.99s & 939.85s & 576.86s & 578.64s & 578.91s & \textbf{206.02s} & 218.22s & 218.06s & 569.57s & 567.55s & 567.86s & 574.09s & 572.32s & 573.75s & 936.62s & 933.18s & 932.34s
\\ \hline
Segment & 72.41s & 72.66s & 72.15s & 36.87s & 36.99s & 36.63s & 37.23s & 37.26s & 37.11s & \textbf{13.02s} & 13.43s & 13.53s & 35.18s & 35.53s & 35.46s & 36.98s & 37.02s & 36.75s & 37.25s & 36.33s & 36.58s
\\ \hline
Sonar & 0.51s & 0.53s & 0.53s & 0.40s & 0.41s & 0.41s & 0.27s & 0.28s & 0.29s & \textbf{0.10s} & 0.11s & 0.13s & 0.28s & 0.34s & 0.41s & 0.28s & 0.31s & 0.33s & 0.40s & 0.40s & 0.41s
\\ \hline
Splice & 37.29s & 34.84s & 34.96s & 30.49s & 30.29s & 29.83s & 18.24s & 18.19s & 18.50s & 6.94s & \textbf{6.91s} & 6.99s & 17.79s & 17.92s & 17.83s & 18.33s & 18.41s & 18.75s & 31.50s & 30.83s & 29.60s
\\ \hline
Vehicle & 10.01s & 9.82s & 9.71s & 4.90s & 4.93s & 4.96s & 4.96s & 5.00s & 5.04s & \textbf{1.77s} & 1.79s & 1.82s & 4.86s & 4.86s & 4.89s & 5.00s & 5.04s & 5.07s & 5.02s & 4.98s & 4.96s
\\ \hline
Wine & 0.44s & 0.44s & 0.44s & 0.22s & 0.23s & 0.23s & 0.22s & 0.23s & 0.23s & \textbf{0.08s} & \textbf{0.08s} & 0.09s & 0.22s & 0.23s & 0.25s & 0.23s & 0.23s & 0.24s & 0.22s & 0.22s & 0.22s
\\
\hhline{-|-|-|-|-|-|-|-|-|-|-|-|-|-|-|-|-|-|-|-|-|-|}
\hhline{-|-|-|-|-|-|-|-|-|-|-|-|-|-|-|-|-|-|-|-|-|-|}
\hhline{-|-|-|-|-|-|-|-|-|-|-|-|-|-|-|-|-|-|-|-|-|-|}
\end{tabular}}
\end{center}
\caption{Average computation time required by the optimization process considering all datasets.}
\label{t.time}
\end{adjustwidth}
\end{table}

Regarding the hypercomplex techniques, such as quaternion and octonion, it is possible to observe that they have an extra computational loop per feature, due to its number of dimensions, e.g., 4 and 8. If the number of selected features is sufficiently smaller to overcome this extra loop, the hypercomplex techniques will achieve a shorter computational time than the conventional ones. For example, in a $100$-features problem, the conventional technique loop lasts for $100$ times, while the quaternion and octonion loops last for $400$ and $800$ times, respectively. If the quaternion-based technique selects $25$ features averagely while the octonion-based one selects $12.5$ features averagely, both will perform a loop that lasts for $100$ times, being comparable to the conventional algorithm.

The results obtained in this study prove the promising use of meta-heuristic optimization techniques when selecting a quasi-optimum subset of features while preserving its performance and discriminative aptitudes. 
 
\subsection{Convergence Analysis}
\label{ss.convergence_analysis}

\begin{figure*}[!ht]
	\centering
	\begin{adjustwidth}{-2cm}{0cm}
	\begin{tabular}{cc}
		\includegraphics[scale=0.25]{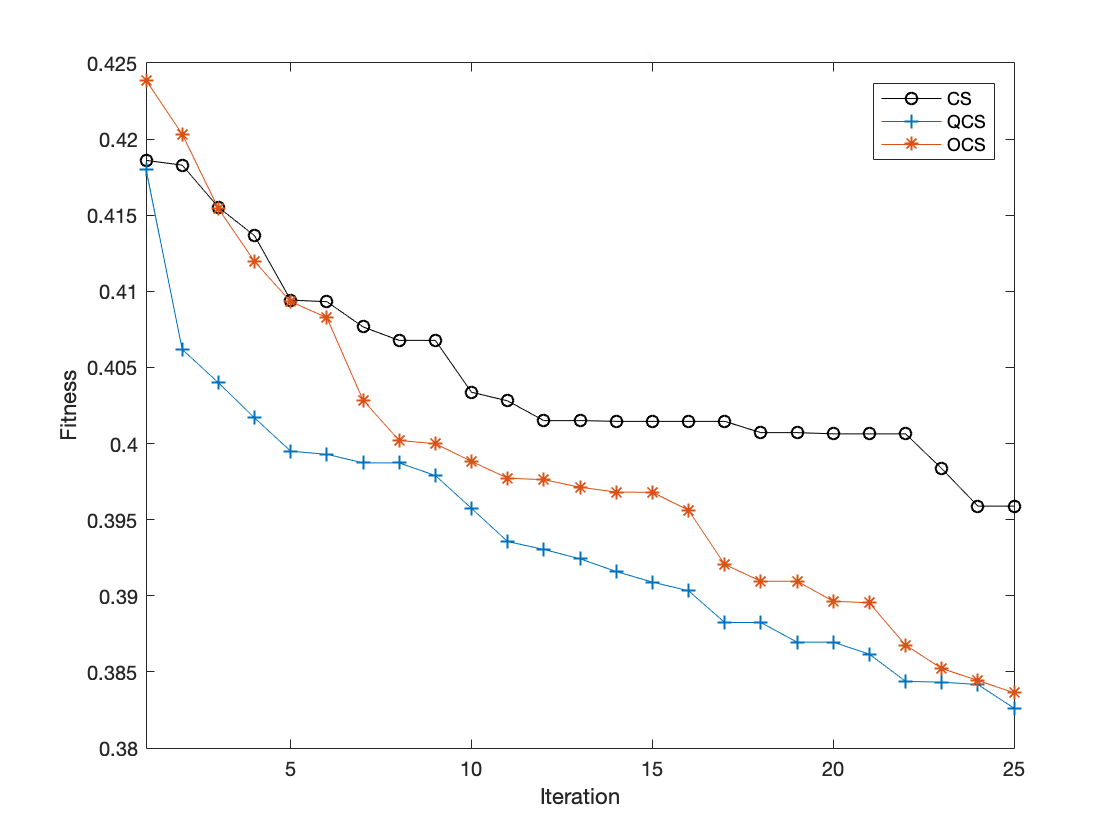} &
		\includegraphics[scale=0.25]{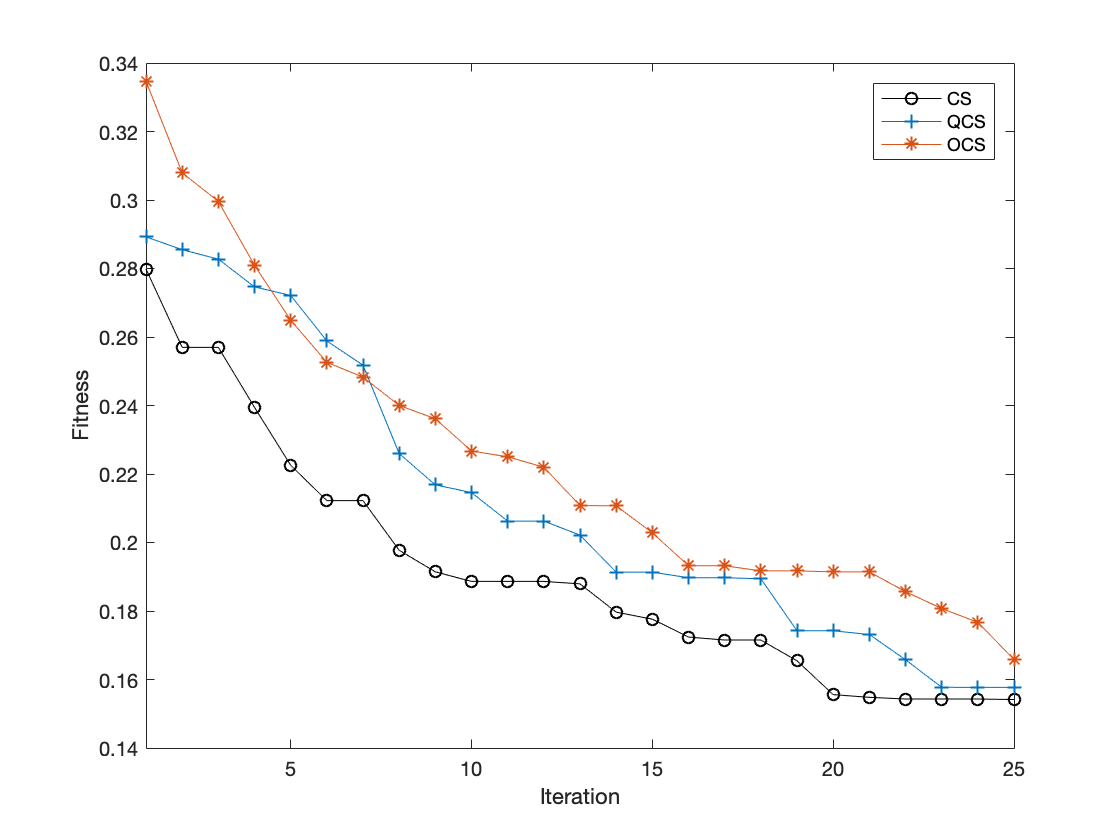}
		\\
		(a) & (b)
		\\
		\includegraphics[scale=0.25]{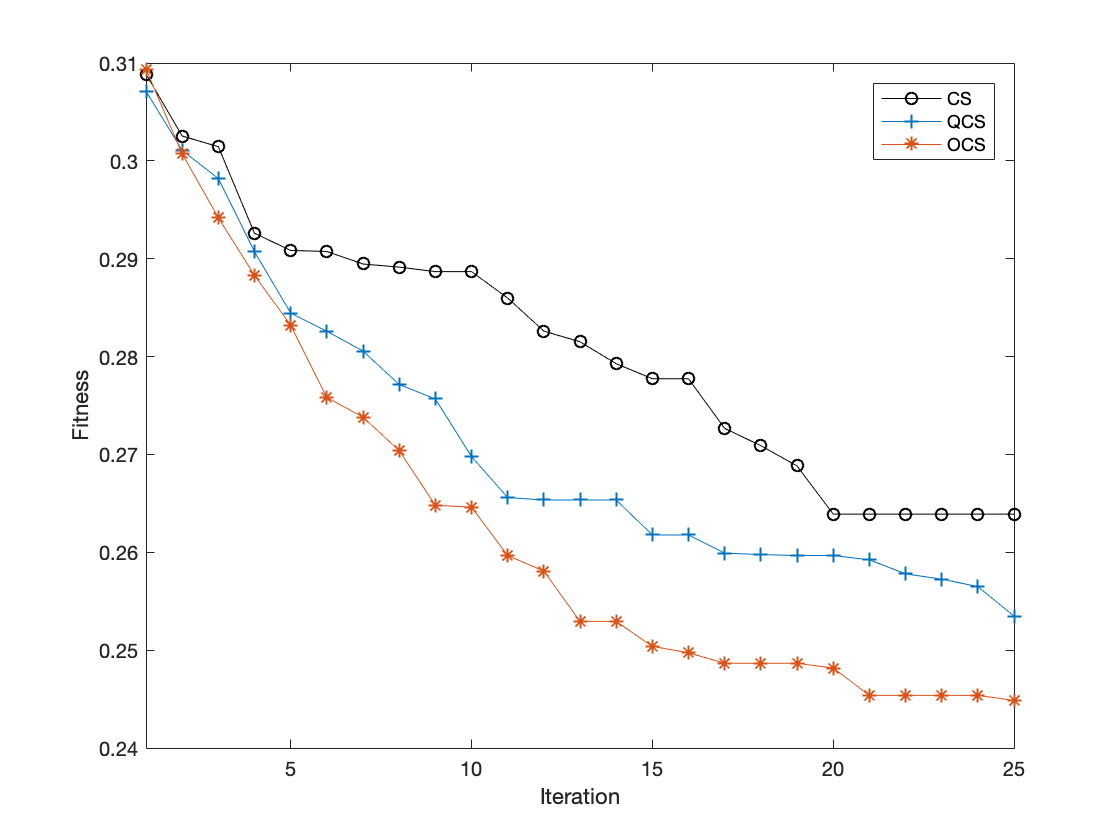} &
		\includegraphics[scale=0.25]{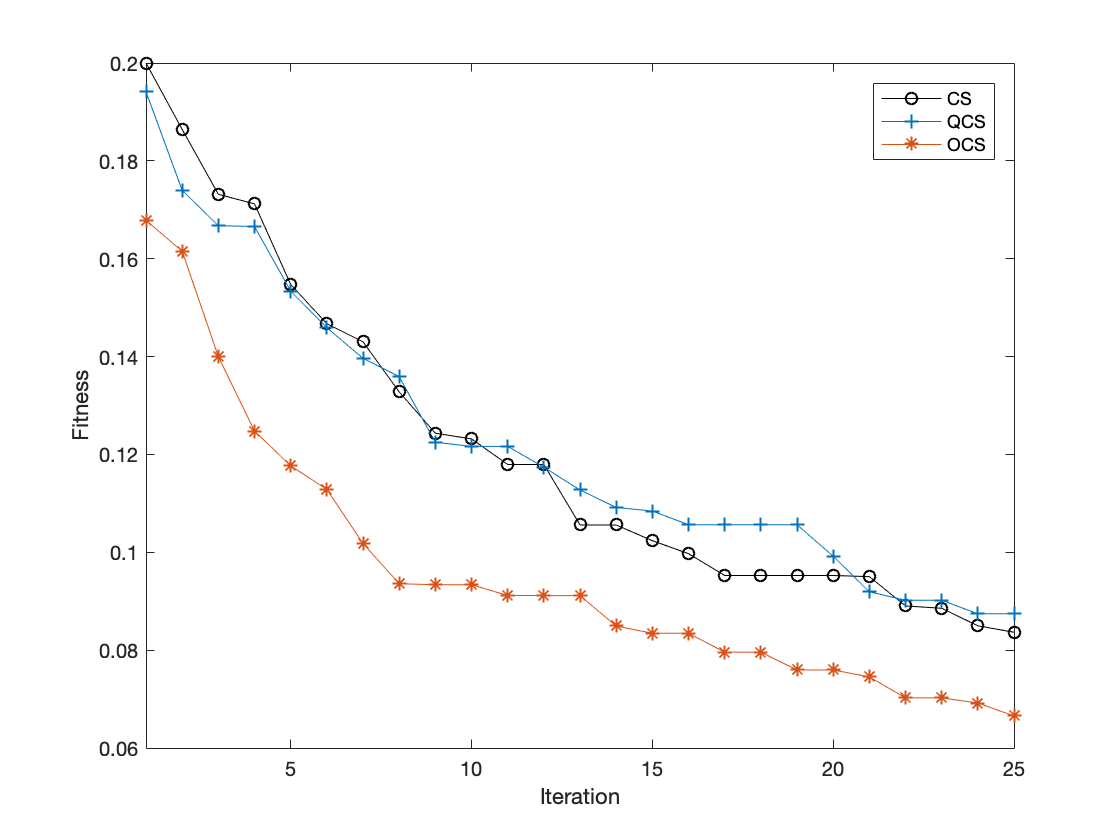}
		\\
		(c) & (d)
	\end{tabular}
	\end{adjustwidth}
	\caption{Iteration x Fitness chart considering CS, QCS and OCS in: (a) DNA, (b) NTL-Commercial, (c) Phishing and (d) Segment datasets.}
	\label{f.cs_convergence}
 \end{figure*}

The convergence curves of CS and its variants obtained for the DNA, NTL-Commercial, Phishing, and Segment datasets are shown in Figure~\ref{f.cs_convergence}.

\begin{sloppypar}
An interesting fact that one can perceive is that hypercomplex-based techniques were able to converge faster and better than the standard version in three out of four datasets (DNA, Phishing, and Segment). Additionally, it is essential to highlight that as hypercomplex-based algorithms use an enhanced version of the search space, i.e., a space with a more substantial amount of possible values, they are capable of better exploring it, thus, leading to better convergence rates and fitness values. Moreover, as OCS encodes a higher-dimensional space, i.e., $8$ dimensions, it was able to achieve the lowest fitness for two datasets (Phishing and Segment), thus showing its exploration capability of the search space.
\end{sloppypar}

 \clearpage

%% file: conclusion.tex
\section{Conclusion}
\label{s.conclusion}

This paper addressed the problem of feature selection through a meta-heuristic optimization approach. A wide range of meta-heuristic techniques was employed in $20$ distinct datasets in order to provide a more thoughtful numerical validation of the proposed computational framework. Additionally, we also present three distinct search spaces for each optimization technique: standard, quaternionic, and octatonic.

In most circumstances, the meta-heuristic techniques were able to outperform the baseline approach (OPF classification over the full-features dataset). In such cases, outperforming means that a singular technique was able to attain higher accuracy than another algorithm, according to the Wilcoxon signed-rank test with 5\% of significance.  Besides, it is possible to highlight that all meta-heuristic techniques were able to diminish a substantial number of the initial datasets' features while maintaining their classification accuracy.

Even though most algorithms were able to reduce the features' space size and obtain statistically similar accuracy within respect to the baseline method, in some cases, they reached a slightly lower accuracy than the original OPF classification. Nevertheless, it should be remarked that in the BASEHOCK dataset, where the baseline classification achieved the best accuracy, all other meta-heuristic techniques could decrease by about 35\% of the number of features while scoring 2-3\% lower accuracy than OPF.

An intriguing fact is that CS was able to obtain the lowest number of features in nearly every dataset, but not necessarily the best accuracy. If one perceives the discrepancy between the best and the worst accuracy considering all datasets (except NTL-based ones) and meta-heuristic techniques, there is not a single one that exceeds the 4.77\% limit. Nonetheless, CS underwent in the NTL datasets (energy theft identification), which are highly unbalanced and have a comparatively small amount of features. As CS obtained the lowest number of features in such a low-dimensional dataset, it is reasonable to mention that is has overfitted the optimization process in an attempt to find the lowest number of possible features. Such a procedure penalized the classifier and, consequently, achieved a not proper accuracy for these particular datasets.

Furthermore, we presented a more in-depth analysis considering CS and its variants, QCS, and OCS, among four distinct datasets that have discrepant data, i.e., datasets with a low amount of features and highly sensitive to feature selection. This analysis provided thoughtful insights regarding the number of selected features per accuracy they were able to achieve, the time they took to perform the optimization process, and their convergence process. Additionally, it is essential to highlight that CS hypercomplex-based approaches took more time than their standard version, while they were able to converge better (to a lower fitness function value) than its na\"ive version.

For future works, we aim at exploring within more depth the hypercomplex mapping function, e.g., norm function. We have high hopes in understanding more the hypercomplex structure, as it seems that one of the central concepts in applying them to feature selection methods lies in transferring values from hypercomplex- to real-valued search spaces.